\documentclass{article} %
\usepackage[final]{colm2026_conference}

\usepackage{fvextra}
\DefineVerbatimEnvironment{wrapverb}{Verbatim}{
  breaklines=true,      %
  breakanywhere=true,   %
  breaksymbolleft={},   %
  breaksymbolsepleft=0pt
}

\usepackage[utf8]{inputenc} %

\usepackage{graphicx}

\usepackage[T1]{fontenc}

\usepackage[hidelinks,colorlinks]{hyperref}       %
\usepackage{url}            %
\usepackage{booktabs}       %
\usepackage{amsfonts}       %
\usepackage{nicefrac}       %
\usepackage{microtype}      %
\usepackage{amssymb}
\usepackage{pifont}

\IfFileExists{headers/config/showoverfull.config}{
	\overfullrule=1cm
}{
}

\usepackage{marginnote}

\usepackage[backgroundcolor=none,linecolor=red,textsize=footnotesize]{todonotes}

\usepackage{etoolbox}

\newbool{includeappendix}
\setbool{includeappendix}{true} %
\IfFileExists{headers/config/noappendix.config}{
	\setbool{includeappendix}{false}
}{}

\newif\ifincludeappendixx
\ifbool{includeappendix}{
	\includeappendixxtrue
}{
	\includeappendixxfalse
}

\usepackage{xr} %
\usepackage{filecontents}

\ifbool{includeappendix}{}{

	\externaldocument{appendix-labels}
}

\newcommand{\eg}{e.g., }
\newcommand{\ie}{i.e., }

\usepackage{acro} %

\DeclareAcronym{cli} {
    short = CLI,
    long = Command Line Interface,
}

\usepackage{listings}

\usepackage{textcomp}

\usepackage{xcolor}

\usepackage[scaled=0.8]{beramono}

\definecolor{ckeyword}{HTML}{7F0055}
\definecolor{ccomment}{HTML}{3F7F5F}
\definecolor{cstring}{HTML}{2A0099}

\lstdefinestyle{numbers}{
	numbers=left,
	framexleftmargin=20pt,
	numberstyle=\tiny,
	firstnumber=auto,
	numbersep=1em,
	xleftmargin=2em
}

\lstdefinestyle{layout}{
	frame=none,
	captionpos=b,
}

\lstdefinestyle{comment-style}{
	morecomment=[l]//,
	morecomment=[s]{/*}{*/},
	commentstyle={\color{ccomment}\itshape},
}

\lstdefinestyle{string-style}{
	morestring=[b]",%
	morestring=[b]',%
	stringstyle={\color{cstring}},
	showstringspaces=false,%
}

\lstdefinestyle{keyword-style}{
	keywordstyle={\ttfamily\bfseries},
	morekeywords={
		function,
		constructor,
		int,
		bool,
		return,
		returns,
		uint
	},
	morekeywords = [2]{},
	keywordstyle = [2]{\text},
	sensitive=true,
}

\lstdefinestyle{input-encoding}{
	inputencoding=utf8,
	extendedchars=true,
	literate=
	{ℝ}{$\reals$}1%
	{→}{$\rightarrow$}1%
	{α}{$\alpha$}1%
	{β}{$\beta$}1%
	{λ}{$\lambda$}1%
	{θ}{$\theta$}1%
	{ϕ}{$\phi$}1%
}

\lstdefinestyle{escaping}{
	moredelim={**[is][\color{blue}]{\%}{\%}},
	escapechar=|,
	mathescape=true
}

\lstdefinestyle{default-style}{
	basicstyle=\fontencoding{T1}\ttfamily\footnotesize,
	style=numbers,
	style=layout,
	style=comment-style,
	style=string-style,
	style=keyword-style,
	style=input-encoding,
	style=escaping,
	tabsize=2,
	upquote=true
}

\lstdefinelanguage{BASIC}{
	language=C++,
	style=default-style
}[keywords,comments,strings]%

\newcommand{\QK}[2][]{%
  \ensuremath{Q#2\_{K%
    \if\relax\detokenize{#1}\relax
    \else \_{#1}%
    \fi}}%
}

\usepackage[utf8]{inputenc} %
\usepackage[T1]{fontenc}    %
\usepackage{algcompatible}
\usepackage{caption}
\usepackage{algpseudocode}
\usepackage{enumerate}
\usepackage{url}            %
\usepackage{booktabs}       %
\usepackage{amsfonts}       %
\usepackage{nicefrac}       %
\usepackage{microtype}      %
\usepackage{xcolor}         %
\usepackage{algpseudocode}
\usepackage{algorithm}
\usepackage{tikz,booktabs,multirow,enumitem,bm}
\usepackage{colortbl}
\usepackage{tabularx}
\usepackage{arydshln}
\usepackage{subcaption}
\usepackage{wrapfig}
\usepackage{tabulary}
\usepackage{amsmath,amsthm,amsfonts}
\usepackage{setspace}
\usepackage{ amssymb }

\usepackage{amsmath,amsfonts,bm}
\usepackage{amsthm}
\usepackage{mathtools}

\def\1{\bm{1}}

\DeclareMathAlphabet{\mathsfit}{\encodingdefault}{\sfdefault}{m}{sl}
\SetMathAlphabet{\mathsfit}{bold}{\encodingdefault}{\sfdefault}{bx}{n}

\definecolor{hyperlinkblue}{HTML}{0000AA}
\hypersetup{citecolor=hyperlinkblue} %

\usepackage[capitalize, noabbrev]{cleveref}

\crefformat{section}{\S#2#1#3}

\crefrangeformat{section}{\S#3#1#4\crefrangeconjunction\S#5#2#6}

\crefmultiformat{section}{\S#2#1#3}{\crefpairconjunction\S#2#1#3}{\crefmiddleconjunction\S#2#1#3}{\creflastconjunction\S#2#1#3}

\newcommand{\crefrangeconjunction}{--}

\crefname{listing}{Lst.}{listings}
\crefname{line}{Lin.}{Lin.}
\crefname{appendix}{App.}{App.}

\newcommand{\appref}[1]{%
	\ifbool{includeappendix}{\cref{#1}}{the appendix}%
}
\newcommand{\Appref}[1]{%
	\ifbool{includeappendix}{\cref{#1}}{The appendix}%
}

\usepackage{lineno}

\definecolor{darkblue}{rgb}{0, 0, 0.5}
\hypersetup{colorlinks=true, citecolor=darkblue, linkcolor=darkblue, urlcolor=darkblue}

\title{Delay, Plateau, or Collapse: Evaluating the Impact of \\ Systematic Verification Error on RLVR}

\author{
  \textbf{Kazuki Egashira$^1$, Mark Vero$^1$, Jasper Dekoninck$^1$, Florian E. Dorner$^{1,2}$, Robin Staab$^1$,} \\
  ~\textbf{Martin Vechev$^1$} \\
  $^1$ ETH Zurich, ~~$^2$ Max Planck Institute for Intelligent Systems, T{\"u}bingen \\
\texttt{kazuki.egashira@inf.ethz.ch} \\
}

\usepackage[most]{tcolorbox}
\usepackage{hhline}
\usepackage{adjustbox}
\newtcolorbox{promptbox}[1][]{
  colback=blue!5!white,
  colframe=blue!75!black,
  boxrule=0.8pt,
  arc=4pt,
  fontupper=\small,
  left=6pt, right=6pt, top=6pt, bottom=6pt,
  fonttitle=\bfseries,
  #1
}

\begin{document}

\ifcolmsubmission
\linenumbers
\fi

\maketitle

\begin{abstract}
Reinforcement Learning with Verifiable Rewards (RLVR) has become a powerful approach for improving the reasoning capabilities of large language models (LLMs). While RLVR is designed for tasks with verifiable ground-truth answers, real-world verifiers (e.g., static code checkers) can introduce errors into the reward signal. Prior analyses have largely treated such errors as random and independent across samples, concluding that errors merely slow training with limited effect on final performance. However, practical verifiers tend to exhibit systematic errors. This introduces a risk of models learning unwanted consistent behavior from a structurally incorrect reward signal.
In this work, we study the impact of such systematic verification errors on RLVR. Through controlled experiments on arithmetic tasks, we show that systematic false negatives lead to similar effects as random noise. On the other hand, systematic false positives can cause a wide range of behaviors from sub-optimal plateaus to performance collapse. Crucially, these outcomes are not determined by the overall error rate but by the specific pattern of introduced errors, making pre-hoc mitigation difficult. Our results show that, in contrast to prior conclusions, realistic verification errors can critically shape RLVR outcomes and that verifier quality has to be understood beyond its sample-level error rate. %

\end{abstract}

\section{Introduction}
\label{sec:introduction}
\vspace{-1mm}

Large language models (LLMs) have advanced rapidly across a wide range of domains and are now able to assist experts with complex tasks such as mathematical reasoning and coding \citep{yang2025qwen3,zeng2025glm}. One of the key techniques behind this progress is Reinforcement Learning with Verifiable Rewards (RLVR) \citep{shao2024deepseekmath}, which applies reinforcement learning using a ground-truth verifier that rewards the model based on whether its output is correct. However, as LLMs are deployed on increasingly complex tasks, designing accurate verifiers becomes challenging \citep{rubricsasrewards} and the reward signal may become \emph{erroneous}. In particular, verifiers may accept incorrect solutions, producing false positives (FP), or reject correct ones, producing false negatives (FN).

\vspace{-1mm}
\paragraph{Impact of imperfect verifiers}
Several works have examined the impact of verification errors on RLVR training \citep{rad2026rate,cai2025reinforcement,lv2025climb}. Their findings suggest that such errors merely delay training progress without changing the final outcome. However, these conclusions largely rely on the assumption that verifier errors are random, and existing experiments typically model noise through random label flipping.
This assumption may not always hold because many practical verifiers, including static analyzers and LLM-as-a-judge, can exhibit systematic and potentially exploitable error patterns \citep{huang2025accuracy,chen2025acereason,zhao2025one}.

\begin{figure}[t]
\centering
\begin{subfigure}[t]{0.58\textwidth}
    \centering
    \input{figures/teaser/overview.tex}
    \caption{Categorization of systematic errors in RLVR.} \label{subfig:teaser_overview}
\end{subfigure}
\hfill
\begin{subfigure}[t]{0.4\textwidth}
    \centering
    \includegraphics[width=\linewidth]{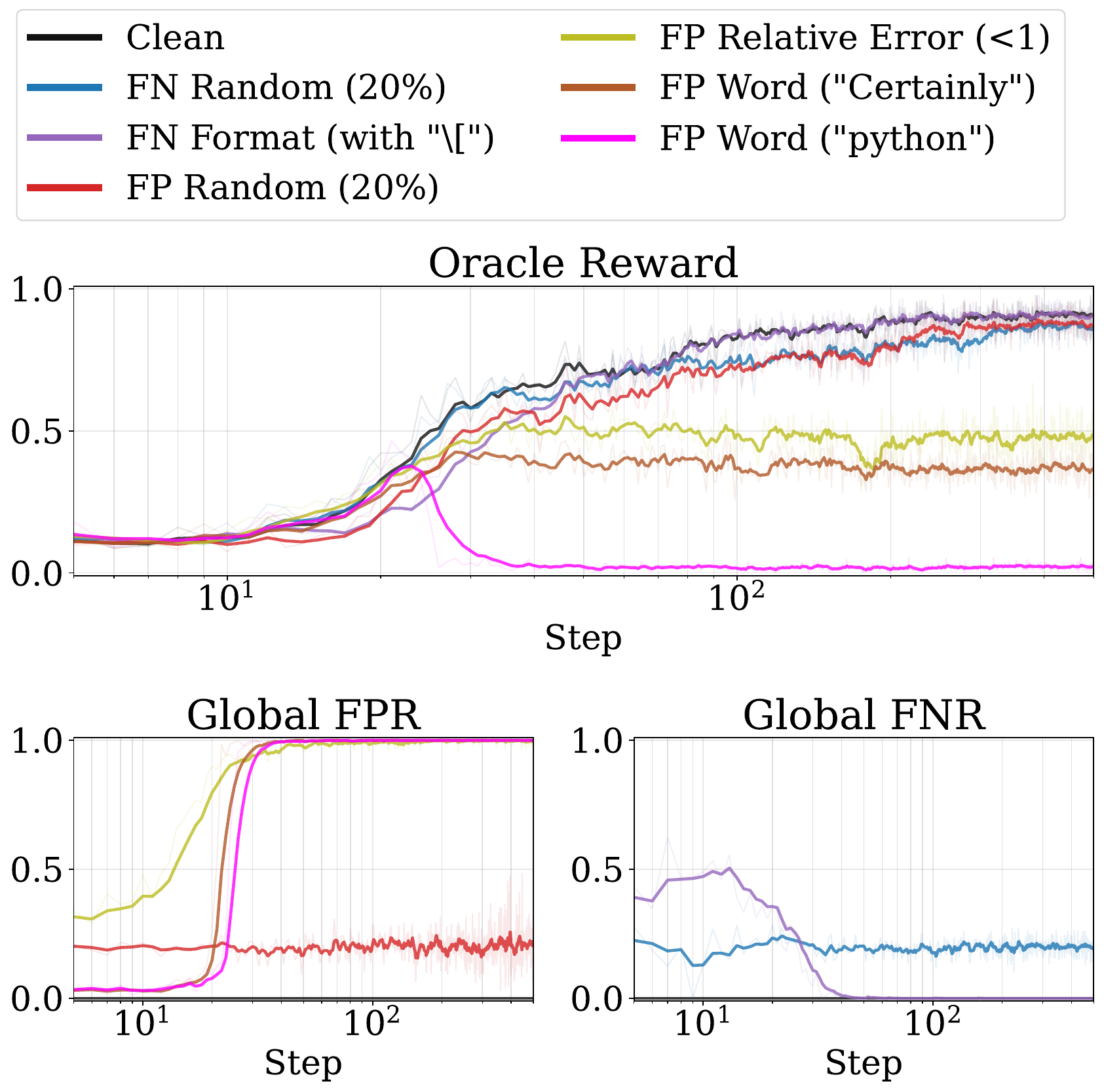}
    \caption{Representative results. (OLMo3-7B)} \label{subfig:teaser_results}
\end{subfigure}
\vspace{-2mm}
\caption{
    Overview of our results. On the left, we categorize systematic errors in RLVR based on their effect on training dynamics, including delayed training, suboptimal plateaus, and training collapse. On the right, we show representative results of these different error types in a controlled experiment using the Reasoning Gym dataset~\citep{stojanovski2025reasoning}.
}
\label{fig:teaser}
\vspace{-1em}
\end{figure}

\vspace{-1mm}
\paragraph{This work: analyzing systematic errors}
In this work, we study the impact of systematic verification errors in RLVR. To do so, we first provide a rigorous definition of systematic verification errors, clearly differentiating them from random noise. Then, in controlled experiments on the arithmetic task from the Reasoning Gym library \citep{stojanovski2025reasoning}, we study how different types of systematic errors affect RLVR training. In particular, we introduce simple, controllable trigger patterns into a ground-truth verifier, resulting in systematic false positives or false negatives. This setup allows us to precisely track all key aspects of the training dynamics, including the error rate, specific error patterns, and the model's behavior.

\vspace{-1mm}
\paragraph{Key findings: delay, plateau, or collapse} As shown in \cref{fig:teaser}, we find that systematic errors can lead to a range of outcomes, including delayed training, suboptimal plateaus, and even complete collapse. Importantly, we find that these behaviors cannot be explained solely by simple sample-level metrics such as false-positive rate (FPR) or false-negative rate (FNR) at the start of training. As shown in \cref{subfig:teaser_results}, total collapse is possible even if the FPR is very low initially, making the impact of verification errors difficult to predict \emph{a priori}.

\paragraph{Implications for RLVR} RLVR is increasingly being applied to complex tasks, where there is an inherent tradeoff between verification cost and accuracy. For example, using a state-of-the-art LLM as a verifier may be prohibitively expensive. This can make smaller models or static analyzers attractive alternatives, even though they may exhibit systematic error patterns. Our findings suggest that the choice of verifier can critically shape RLVR outcomes, and that understanding a verifier's specific error patterns is essential for anticipating their effects. Importantly, we also find that averaged \emph{a priori} error rates are insufficient to predict whether a given verifier is reliable enough to attain reasonable performance.
For instance, as we show in~\cref{subsec:relative_error_fp}, collapse can arise under error patterns that are asymmetric around the ground truth, even when their symmetric counterpart, despite having a strictly higher error rate, only leads to a plateau. This makes it crucial to analyze and improve verifier quality in ways that go beyond aggregate error rates.

\paragraph{Key contributions:}

\begin{itemize}[itemsep=0em,parsep=0.5em]
    \vspace{-2mm}
    \item We analyze \textit{systematic} verification errors in RLVR
    \footnote{Code available at: \url{ https://github.com/eth-sri/llm-verifier-noise}}
    and categorize them according to their effect on training dynamics (\cref{sec:systematic_noise}).
    \item We design controlled experiments to analyze the impact of noisy verifiers, finding delayed training, suboptimal plateaus, and even training collapse (\cref{subsec:main_result}).
    \item We analyze, discuss, and provide a practical outlook for the impact of verification errors (\cref{subsec:relative_error_fp,subsec:word_based_fp,subsec:language_fn}). Our results indicate that an initial error rate of the verifier does not fully explain the training outcome.
\end{itemize}

\section{Related Work}
\label{sec:related_work}
In this section, we briefly review the relevant literature on RLVR and on the impact of verification errors during RLVR training.

\paragraph{RLVR}
RLVR is a widely used post-training technique for improving the reasoning capabilities of LLMs \citep{guo2025deepseek,shao2024deepseekmath,team2025kimi,qwen3technicalreport,zeng2025glm}. It has been applied extensively in domains where verification is relatively simple, including mathematics based on final answers and formal proofs \citep{godelproverv2,guo2025deepseek} and code generation \citep{rltf}. More recently, it has also been extended to broader settings, with rewards based on rubrics  \citep{rubricsasrewards} or
 semantic similarity in code \citep{swerl}. RLVR is often implemented using Group Relative Policy Optimization (GRPO) \citep{shao2024deepseekmath} or related variants \citep{yu2025dapo,liu2025understanding,gao2025soft}.
In these methods, the model generates multiple rollouts for each problem, and a \textit{verifier} assigns a scalar \textit{reward} to each rollout. The rewards are then compared within each problem-specific group to compute an \textit{advantage}, which serves as the learning signal.

\vspace{-1mm}
\paragraph{Verification errors}
In practice, RLVR rewards are often erroneous because verifiers accept incorrect solutions or reject correct ones.
For example, unit tests rarely provide complete coverage \citep{liu2023your}, and rule-based verifiers can reject correct answers because of formatting issues \citep{xu2025tinyv}.
LLM-based verifiers are also increasingly used, but they often exhibit high false-positive rates \citep{raina2024llm,zhou2025variation,zhao2025one}.
This has motivated growing interest in measuring verifier quality.
In particular, \citet{li2026verifybench,yan2025verifybench} propose benchmarks for evaluating verifier accuracy across domains.

\vspace{-1mm}
\paragraph{The impact of random noise on RLVR}
Several works directly analyze the effect of random verification errors on RLVR. For example, \citet{lv2025climb} argue that RLVR tolerates up to $40\%$ randomly flipped labels with little performance loss. In contrast, \citet{cai2025reinforcement} report a stronger negative effect, although they find that the lost performance can be recovered with an algorithmic correction that is effectively a change in learning rate. \citet{rad2026rate} complement these empirical results with a theoretical analysis based on Youden's $J := \mathrm{TPR} - \mathrm{FPR}$: When $J<0$, training collapses, whereas perfect learning is possible when $J>0$. Overall, these works show that aggregate noise rates can explain RLVR behavior when verification errors are modeled as random label flips. Our work instead studies \emph{systematic} errors that the model may learn to trigger or avoid during training.

\vspace{-1mm}
\paragraph{The impact of systematic errors on RLVR}
A small body of work moves beyond random-noise assumptions and shows that imperfect verification can harm RLVR in specific settings. For instance, \citet{huang2025accuracy} study both rule-based and LLM-based verifiers for final-answer mathematical problems. They find that LLM-based verifiers achieve higher accuracy on a fixed dataset, but are also more susceptible to reward hacking, leading to worse post-RLVR performance. Similarly, \citet{chen2025acereason} show that introducing verification errors in code RL harms performance, and \citet{yan2025verifybench} find that verifier accuracy is correlated with RLVR performance on math for LLM-based verifiers. Lastly, \citet{zhu2026noisy} find that while randomizing ground-truth answers causes RLVR performance to fully collapse, replacing ground truth with wrong samples from another LLM leads to diminished performance gains. These works show that verifier failures can harm RLVR in specific settings, but they do not isolate how different systematic error patterns induce delayed learning, plateaus, or collapse under controlled conditions.

\vspace{-1mm}
\paragraph{Reward hacking} Failures in reinforcement learning due to systematic errors, especially false positives, can be seen as a special case of reward hacking.
Reward hacking is a well-known phenomenon, where agents exploit loopholes in the reward function to obtain high rewards without solving the intended task \citep{rewardhacking1,rewardhacking2}. In particular, \citet{gao2023scaling} derive scaling laws for reward hacking in RLHF with continuous reward models, while  \citet{dorner2025roc} characterize the impact of verification errors on resampling-based test-time scaling. However, no prior work has studied the connection between systematic verification errors and RLVR, which is the focus of our work.

\section{Characterizing Training Dynamics with Imperfect Verifiers}
\label{sec:systematic_noise}
In this section, we define systematic errors in the context of RLVR and characterize the downstream training dynamics that it can induce. We first introduce the necessary notation and definitions, and then describe the different training dynamics.

\paragraph{Rewards and verifiers}
A verifier is a function $V$ that takes a query $x$ and a model output $y$ and produces a reward signal $r = V(x,y)$. In this work, we restrict ourselves to binary rewards $r \in \{0,1\}$. With $V^*$, we denote the hypothetical ground-truth verifier that produces the correct reward signal $r^*$ for any input-output pair. However, in training $V^*$ is often too expensive to use or technically infeasible and therefore replaced by a cheaper but imperfect verifier $V$. In RLVR, a model $M$ is then trained to maximize the expected reward obtained using the verifier. In practice, this objective is optimized with approximate reinforcement learning algorithms such as GRPO \citep{shao2024deepseekmath}. These algorithms typically involve sampling multiple rollouts for each problem and train the model using advantages, computed as $A_i = (r_i - \mu(r))/\sigma(r)$, where $r_i$ is the reward assigned to a rollout with average reward $\mu(r)$ and standard deviation $\sigma(r)$ computed across rollouts for the same query.

\paragraph{Performance metrics}
During training, we track several metrics to analyze training dynamics. Our main metric is the \textit{oracle reward} $R^*(M)$, defined as the average reward assigned by the ground-truth verifier $V^*$ to the model's outputs.
This metric reflects the model's true task performance, independent of verification errors. We also track the \textit{verifier reward} $R(M)$, defined as the average reward assigned by the imperfect verifier $V$ to the model's outputs. This is the reward signal the model actually optimizes during training. Finally, we track the false positive rate (FPR) and false negative rate (FNR) of the verifier $V$ relative to the ground-truth verifier $V^*$. Importantly, as most of these essential metrics require access to the ground-truth verifier $V^*$, we restrict ourselves to settings where $V^*$ is tractable.

\paragraph{Verification errors}
We distinguish between two types of verification errors. First, \emph{random noise} is independent of the query $x$ and model output $y$, conditional on the ground truth reward $V^*(x,y)$
and is typically modeled by flipping the reward with a fixed probability. In particular, its FPR and FNR remain constant throughout training. In contrast, \emph{systematic errors} are a function of both the query and model output.
For example, the verifier may assign a false positive reward when the output is close to the ground-truth answer, or when it contains a particular keyword. In such cases, the FPR and FNR can change throughout training because the model may learn to trigger or avoid these systematic verification errors.

\paragraph{Characterization of training dynamics}
In our experiments, we observe four distinct training dynamics, which we characterize based on their reward curves:
\begin{itemize}[leftmargin=2em]
    \item \textbf{Ideal}: The verifier $V$ is effectively the same as the ground-truth verifier $V^*$, meaning that it produces no systematic errors. This is the ideal regime and occurs when both FPR and FNR remain (close to) $0$ throughout training.
    \item \textbf{Delayed}: Delayed training is characterized by an oracle reward that stays consistently below the reward obtained in ideal training, but eventually reaches a similar final value. Prior work \citep{rad2026rate,cai2025reinforcement,lv2025climb} has shown that this behavior occurs under random noise. As we will show, it can also occur when the verifier produces only systematic false negatives, since the model can learn to avoid outputs that trigger those false negatives while still improving task performance. %
    \item \textbf{Plateau}: Plateauing is characterized by an oracle reward that increases throughout training, but converges to a suboptimal value that is below the reward obtained in ideal training. This often occurs when the verifier produces systematic false positives for easy-to-learn triggers that do not have negative interference with the true reward. In this case, the observed reward quickly saturates, preventing the model from learning behavior that actually improves task performance, because it receives maximal reward regardless of whether it solves the task.
    \item \textbf{Collapse}: Collapse is characterized by an oracle reward that eventually decreases during training. Typically, the final performance of the associated model is very low. When this performance is close to random guessing, we refer to it as \textit{complete collapse}. This can occur when the verifier produces systematic false positives for triggers that are easy to learn and conflict with the true reward. In this case, the model learns to produce outputs that trigger false positives, which leads to a negative learning signal for behavior that would actually improve task performance, causing the model to unlearn such behavior and eventually collapse.
\end{itemize}

\section{Experiments}
\label{sec:experiments}
In this section, we first describe the setup  (\cref{subsec:setting}) and then demonstrate that all training dynamics from \cref{sec:systematic_noise} can occur in practice (\cref{subsec:main_result}). Next, we take a deeper look at the error patterns themselves, analyzing their training dynamics in more detail (\cref{subsec:relative_error_fp,subsec:word_based_fp,subsec:language_fn}).
Additional results and analyses are provided in \cref{appsec:more_results}.

\subsection{Experimental Setup}
\label{subsec:setting}

We discuss the main experimental setup here, and defer additional details to \cref{app:experimental_details}.

\paragraph{Evaluation environment}
To study the effects of systematic verification errors rigorously, we aim to use an environment in which the verifier's error patterns can be precisely controlled and their evolution tracked throughout training. Additionally, in order to allow accurate differentiation between the different training dynamics, clean training needs to improve performance significantly. We therefore focus on the decimal chain sum task from Reasoning Gym \citep{stojanovski2025reasoning}, where each example is a multi-term floating-point arithmetic problem, such as \texttt{"332.419 - 993.538 + 740.756 = ?"}.

\paragraph{Training}
We use TRL \citep{vonwerra2020trl} as the training framework and focus on its default DAPO-style loss normalization \citep{yu2025dapo}, which provides a slight improvement over the original GRPO \citep{shao2024deepseekmath}.
Results for Dr.~GRPO \citep{liu2025understanding} and SAPO \citep{gao2025soft} are similar and are reported in \cref{appsubsec:different_algorithms}.
For training, we consider two models from different model families: Qwen3-1.7B-Base~\citep{qwen3technicalreport} and OLMo3-7B~\citep{olmo2025olmo}. We also provide results initialized from an instruction-tuned model, Qwen2.5-1.5B-Instruct, in \cref{appsubsec:instruct}.

\paragraph{Estimating metric values} We generate 256 rollouts (4 per query) in each step and directly use them to compute unbiased estimates of the reward, FPR, and FNR. This avoids generating extra outputs solely for evaluation, which would substantially reduce the number of training steps possible under a fixed compute budget. Note that we always train for only one epoch to avoid contamination of these estimates.

\paragraph{Introduced error patterns} We separately introduce both random noise and various types of systematic error into the verifier. In particular, we consider the following  patterns:
\begin{itemize}[leftmargin=2em]
    \item \textbf{Random noise}: Randomly flips the reward with a fixed probability for either positives or negatives. We consider FPR and FNR  at levels of $20\%$ and $50\%$, respectively.
    \item \textbf{Format-based false negatives}: Incorrectly assigns reward 0 to correct answers that appear in a given output format. In particular, we consider the token \texttt{"\textbackslash["} as a trigger, since it is a common formatting pattern observed under clean-verifier training.
    \item \textbf{Language-based false negatives}: Incorrectly assigns reward 0 to correct answers that are in English, which is an extreme example of systematic false negatives.
    \item \textbf{Relative error-based false positives}: Incorrectly assigns reward 1 to incorrect answers whose relative error from the ground-truth answer falls below a fixed threshold.
    \item \textbf{Word-based false positives}: Incorrectly assigns reward 1 to any output that contains a predefined keyword, regardless of correctness.
\end{itemize}
A detailed description of each error pattern is provided in \cref{appsubsec:error_design}. In \cref{appsubsec:more_error_type}, we also study length-based false positives.

\subsection{Training Dynamics under Systematic Verification Errors}
\label{subsec:main_result}

\begin{figure}[t]
\centering
\setlength{\tabcolsep}{3pt}
\begin{tabular}{cc}
\textbf{OLMo3-7B} & \textbf{Qwen3-1.7B-Base} \\
\midrule

\multicolumn{2}{c}{\textit{FNR}} \\
\multicolumn{2}{c}{\includegraphics[width=0.7\textwidth]{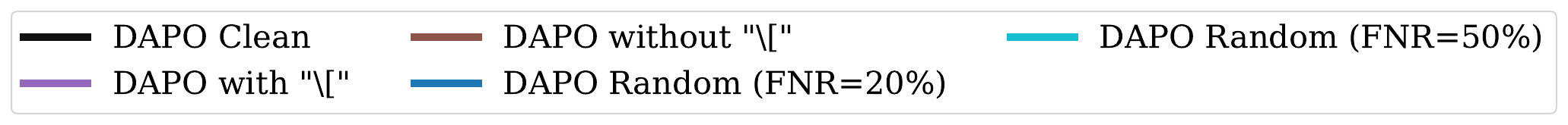}} \\
\includegraphics[width=0.48\textwidth]{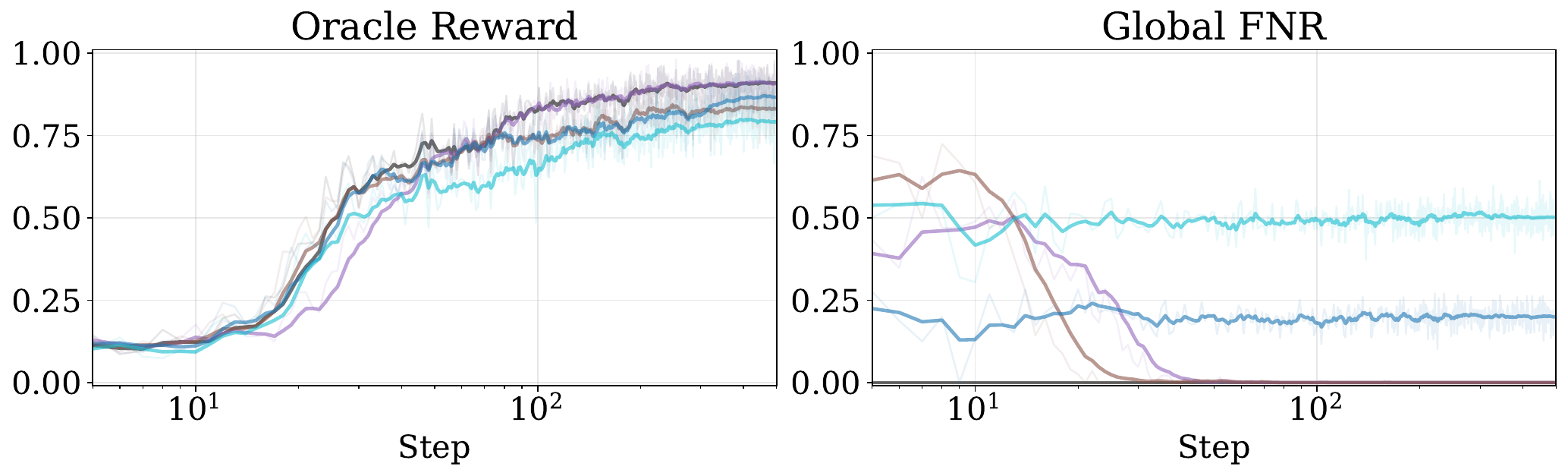} &
\includegraphics[width=0.48\textwidth]{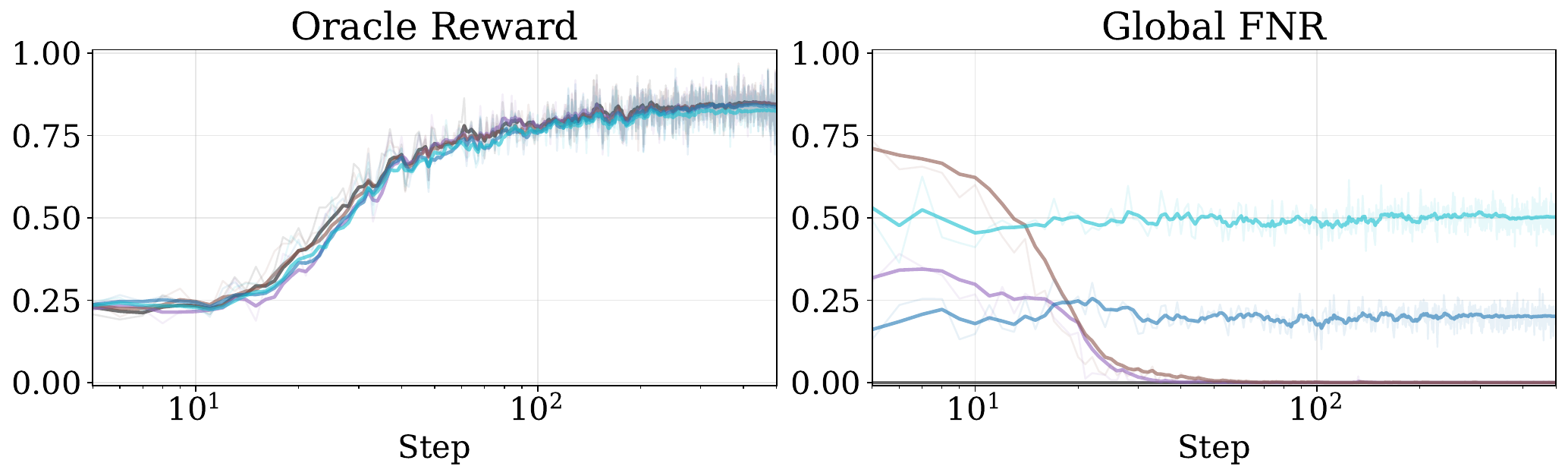} \\

\multicolumn{2}{c}{\textit{FPR}} \\
\multicolumn{2}{c}{\includegraphics[width=0.7\textwidth]{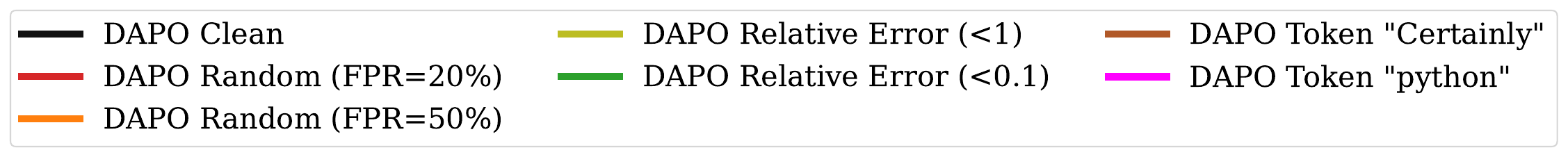}} \\
\includegraphics[width=0.48\textwidth]{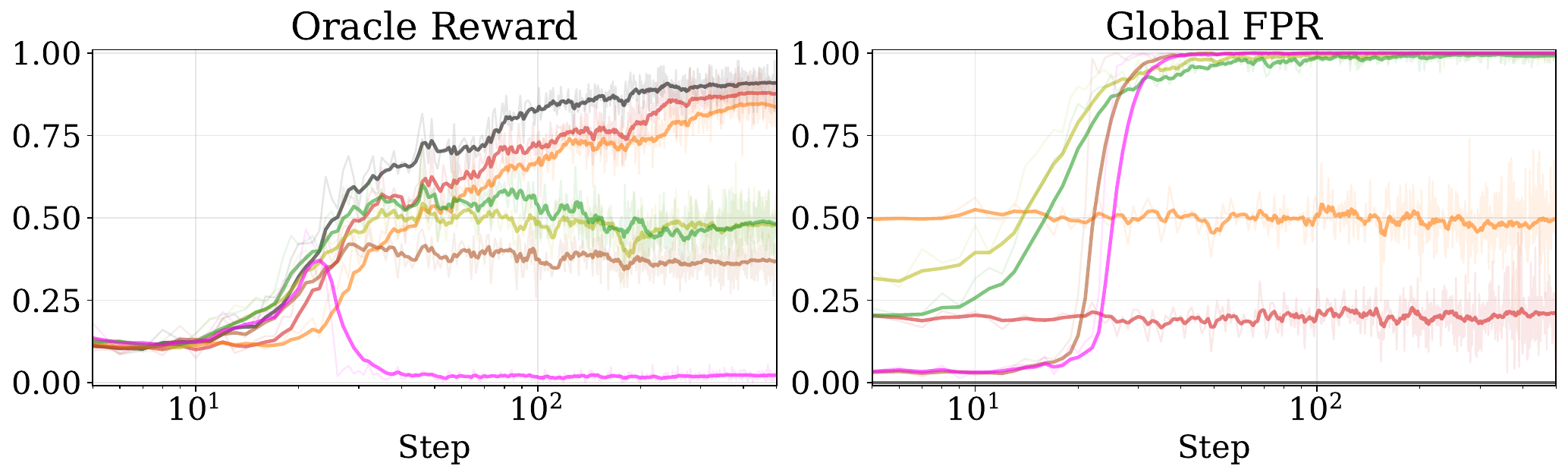} &
\includegraphics[width=0.48\textwidth]{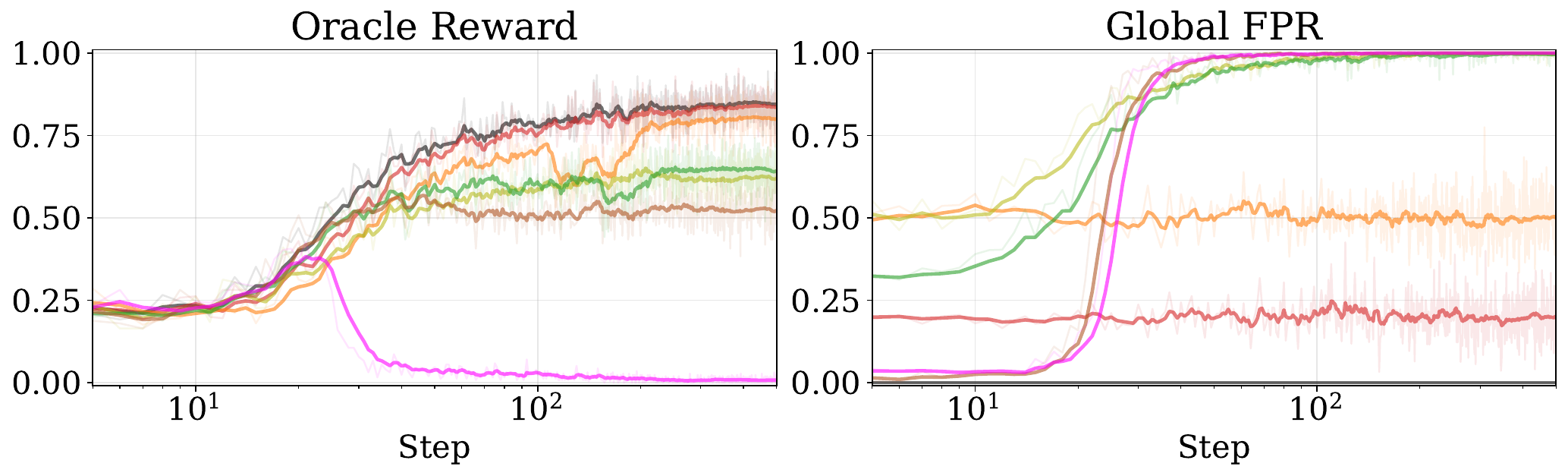} \\

\end{tabular}
\vspace{-3mm}
\caption{
    \textbf{Main results across two models (left: OLMo, right: Qwen).}
    We observe the following trends: (i) random noise and systematic FNR lead to delayed training but similar final performance, while systematic FPR leads to plateauing or collapsing behaviors, depending on the trigger used. We provide results with Dr. GRPO and SAPO in~\Cref{appfig:all_algorithms}.
}
\label{fig:main_result}
\end{figure}

As shown in \cref{fig:main_result}, all training dynamics described in \cref{sec:systematic_noise} can occur in practice under systematic error patterns. We discuss the results by error type below, and defer a deeper analysis of the error patterns to subsequent sections.

\paragraph{Delayed behavior} Confirming prior work \citep{rad2026rate}, we find that moderate levels of random noise delay training, but do not significantly affect final performance.
Similarly, format-based FN also induces delayed training: the model first unlearns behavior that triggers false negatives and then receives valid training signal, eventually reaching a similar reward as clean training. In \cref{appsec:more_results}, we report results for a systematic but essentially unexploitable FP pattern, where the training curve also follows a delayed trajectory.

\paragraph{Plateau and collapse} In contrast, systematic FPs lead to qualitatively different dynamics. First, with relative error-based FP, training consistently plateaus at a suboptimal reward level. The FPR rises to nearly 1 by about step 50, after which almost every rollout receives a positive reward, leaving little useful learning signal and causing training to stall. Training does not collapse, however, since approximate correctness does not undo the true reward signal. With word-based FP, the training dynamics depend on the keyword used. With \texttt{"Certainly"}, training plateaus at a suboptimal level, while \texttt{"python"} has training collapse to near-zero reward.
We investigate the reasons behind these differences in the next section.

\subsection{Conditional Advantage and Initial Trigger Frequency}\label{subsec:word_based_fp}
We now investigate which factors determine whether a given error pattern leads to collapse or plateau. To do so, we first qualitatively examine the behaviors learned under two word-based false positive settings from \cref{subsec:main_result}, which produce plateau and collapse, respectively. We then introduce the notion of \textbf{conditional advantage} to quantitatively characterize the relationship between the behavior induced by a trigger pattern and the resulting dynamics.

\paragraph{Behavioral patterns} To explain the different training outcomes for the two keywords in \cref{subsec:main_result}, we qualitatively analyze model outputs under the two triggers, with an example shown in \cref{fig:prompt_compare}. We find that models do not hack the verifier by simply inserting the trigger word into an otherwise unchanged output. Instead, they learn a distinct \emph{behavioral pattern} associated with the trigger. With \texttt{"Certainly"}, the model produces mostly correct reasoning, simply inserting the trigger word at the start. In contrast, with \texttt{"python"}, the model learns to emit Python-like code and places an arbitrary number in an \texttt{output} block. Therefore, training reinforces an unhelpful behavior and the model collapses.

The results in \cref{subsec:main_result} suggest that the different training dynamics under the two word-based FP settings are driven by the different behaviors associated with the two trigger words. We hypothesize that the training outcome depends on whether the induced behavior is positively or negatively associated with true task success. To investigate this hypothesis, we introduce the notion of \emph{conditional advantage}.

\begin{figure*}[t]
\centering
\begin{subfigure}[t]{0.32\textwidth}
	\centering
	\input{prompts/clean.tex}
\end{subfigure}
\hfill
\begin{subfigure}[t]{0.32\textwidth}
	\centering
	\input{prompts/certainly.tex}
\end{subfigure}
\hfill
\begin{subfigure}[t]{0.32\textwidth}
	\centering
	\input{prompts/python.tex}
\end{subfigure}
\caption{
	Output examples under different word-based FPs. The clean and \texttt{Certainly}-triggered runs produce correct arithmetic reasoning, while the \texttt{python}-triggered fails by using incorrect code-based output.
}
\label{fig:prompt_compare}
\end{figure*}

\begin{wrapfigure}{r}{0.51\textwidth}
    \centering
    \vspace{-1em}
    \includegraphics[width=0.85\linewidth]{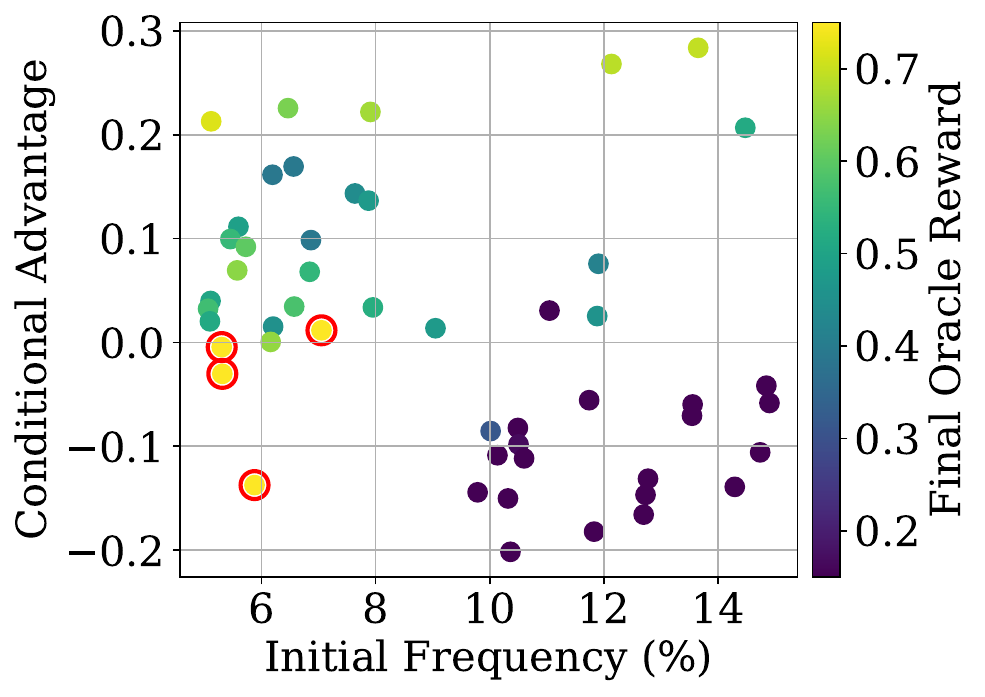}
    \vspace{-1em}
    \caption{
        Distribution of tri-gram and final performance (OLMo).
        The trigrams that remained below 50\% FPR are highlighted in red.
    }
    \label{fig:token_dot}
    \vspace{-2em}
\end{wrapfigure}

\paragraph{Conditional advantage} The conditional advantage $C(t)$ of an error pattern $t$ measures how well the behavior associated with $t$ aligns with the oracle reward. To compute it, we first compute the advantage $A^*(y_i)$ of each output $y_i$ based on the ground-truth verifier $V^*$. We then identify the subset of outputs that contain the trigger pattern $t$, denoted as $S_t$. Finally, we average the advantage over outputs in $S_t$ to obtain the conditional advantage:
\begin{align*}
    C(t) = \frac{1}{|S_t|} \sum_{y_i \in S_t} A^*(y_i).
\end{align*}

Intuitively, conditional advantage measures whether outputs containing the pattern tend to be better or worse than average under the oracle reward: it is maximized if all outputs in $S_t$ are correct and all outputs outside $S_t$ are incorrect, and minimized in the opposite case.

\paragraph{Experimental setup} Now, we repeat the word-based FP experiment with a variety of different trigrams as triggers, and analyze the relationship between the initial frequency of the trigram, its conditional advantage, and the training outcome. We opt for trigrams rather than single tokens for this analysis because they are more behavior-specific and show a more consistent trend than single tokens.

\paragraph{Results}
In \cref{fig:token_dot}, we plot the relationship between a pattern's initial frequency, its conditional advantage, and the final reward. The results support our hypothesis. When a  trigger pattern is frequent at initialization, its effect depends on its conditional advantage. Frequent patterns with negative conditional advantage collapse because rewarding them reinforces behavior that is actively misaligned with the task. By contrast, frequent patterns with positive conditional advantage tend to produce a plateau rather than a collapse. Although rewarding these triggers mis-specifies the objective, the induced behavior remains partially aligned with the task.

When a pattern is rare at initialization, its influence is usually weaker, but conditional advantage still matters.
Rare patterns with positive conditional advantage can still be amplified because they align with the oracle reward and are learned alongside correct behavior. As such, they become more frequent during training and the oracle reward plateaus. By contrast, rare patterns with negative conditional advantage become less frequent during training because they are not reinforced by the oracle reward, so they tend to move further left in the plot, and therefore do not affect training.

Overall, these results suggest that the effect of word-based, or more broadly \emph{behavior-based}, false positives is governed mainly by two factors: how often the trigger-associated behavior appears at initialization and how well that behavior aligns with the oracle reward.

\subsection{Asymmetric Verification Errors Induce Collapse}\label{subsec:relative_error_fp}

One property of relative error-based false positives is that the false-positive region is symmetric around the ground-truth answer.
We therefore ask what happens when this region becomes asymmetric.
Concretely, we cut the false-positive interval in half and place it entirely on one side of the ground truth, either above or below it.

\begin{figure}[t]
\centering

\begin{minipage}[t]{0.49\textwidth}
    \centering
    \vspace{0pt}
    \begin{subfigure}{\linewidth}
        \centering
        \includegraphics[width=0.7\linewidth]{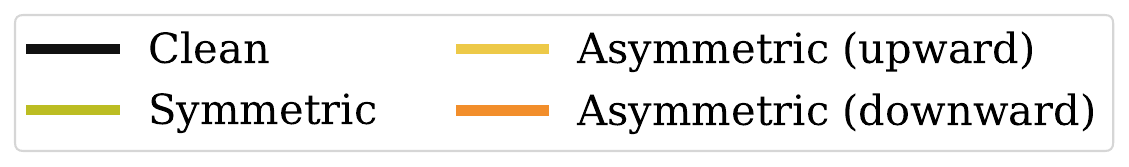}
        \includegraphics[width=\linewidth]{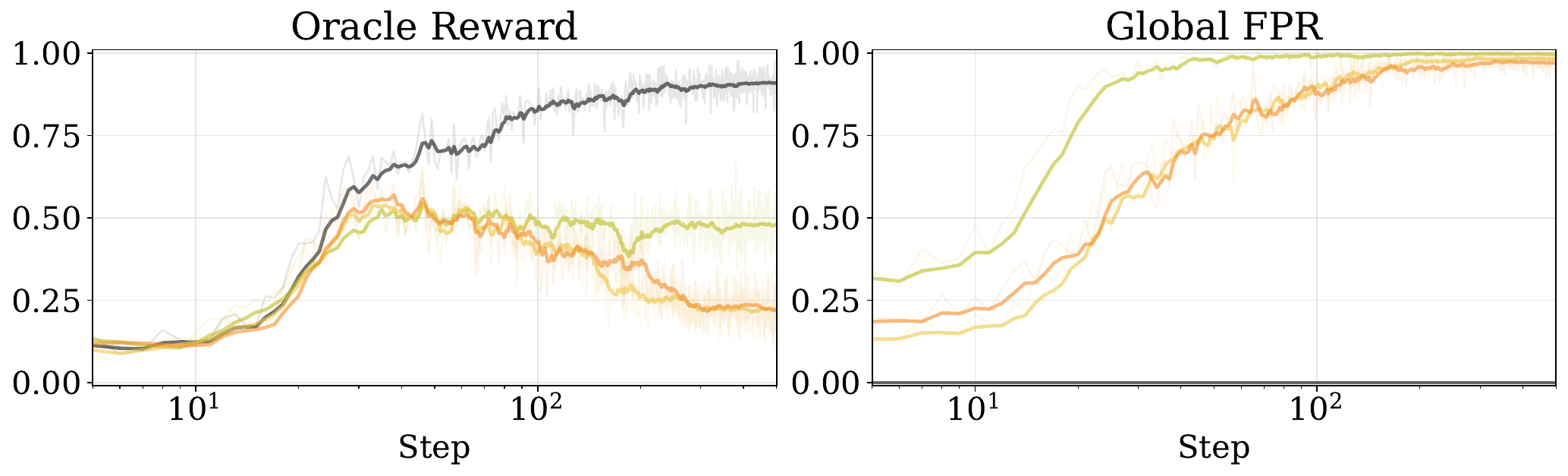}
        \caption{OLMo}
        \label{fig:asymmetric_relative_error_olmo}
    \end{subfigure}

    \begin{subfigure}{\linewidth}
        \centering
        \includegraphics[width=\linewidth]{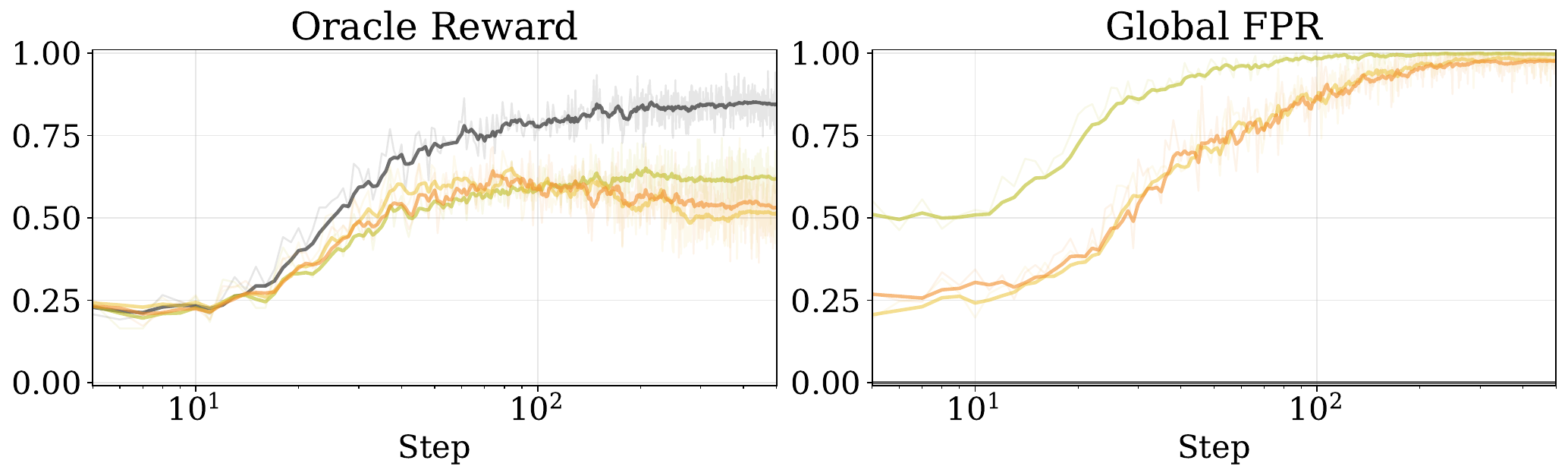}
        \caption{Qwen}
        \label{fig:asymmetric_relative_error_qwen}
    \end{subfigure}

    \begin{minipage}{0.98\linewidth}
        \caption{
            Asymmetric relative error results.
            The training outcomes for asymmetric intervals are worse than the symmetric interval, although the intervals are strictly tighter.
        }
        \label{fig:asymmetric_relative_error}
    \end{minipage}
\end{minipage}
\hfill
\begin{minipage}[t]{0.49\textwidth}
    \centering
    \vspace{0pt}
    \begin{subfigure}{\linewidth}
        \centering
        \includegraphics[width=0.8\linewidth]{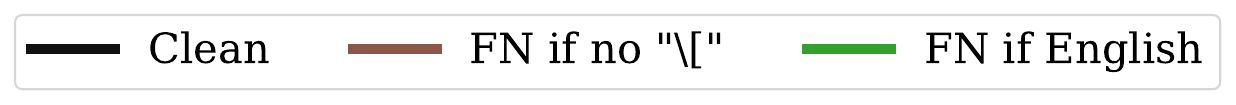}
    \end{subfigure}

    \vspace{0.6em}

    \begin{subfigure}{\linewidth}
        \centering
        \includegraphics[width=\linewidth]{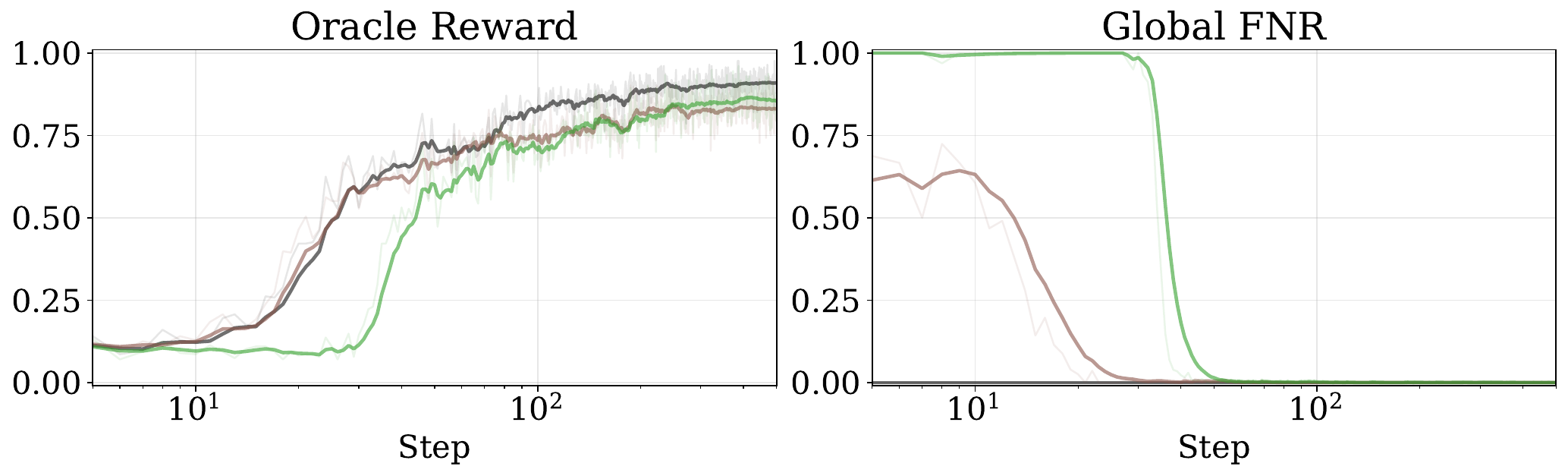}
        \caption{OLMo}
        \label{fig:language_fn_olmo}
    \end{subfigure}

    \begin{subfigure}{\linewidth}
        \centering
        \includegraphics[width=\linewidth]{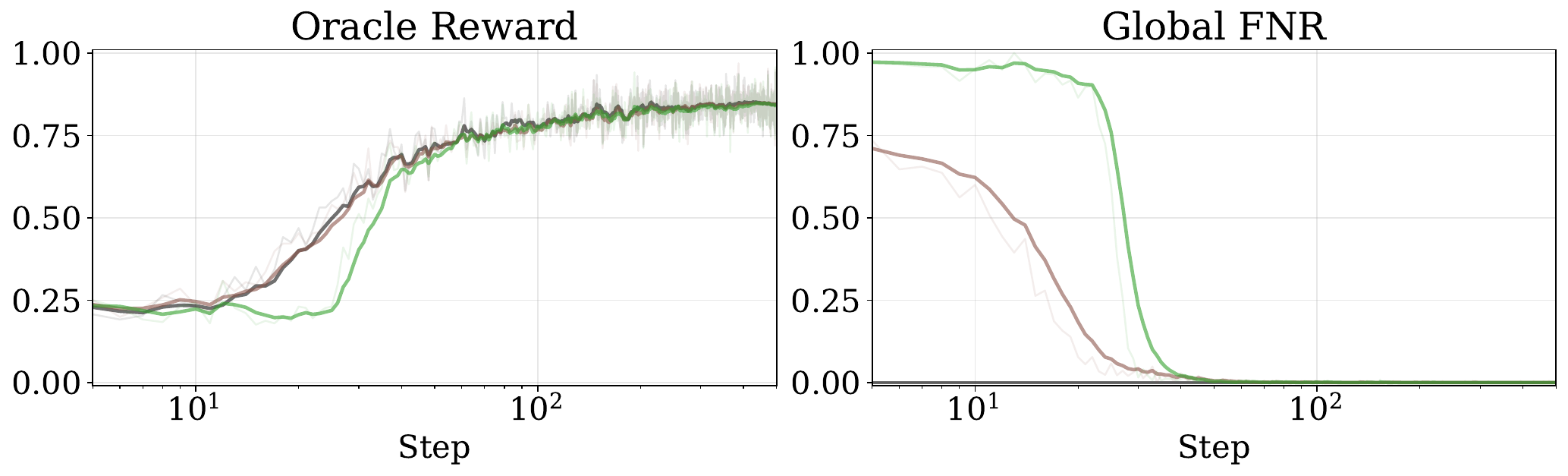}
        \caption{Qwen}
        \label{fig:language_fn_qwen}
    \end{subfigure}

    \begin{minipage}{0.98\linewidth}
        \caption{
            Language-based FN results. For comparison, we also include a format FN, where the verifier gives a false negative if the answer does not contain "\textbackslash[".
        }
        \label{fig:language_fn}
    \end{minipage}
\end{minipage}

\end{figure}

\paragraph{Results}
The results, shown in \cref{fig:asymmetric_relative_error}, demonstrate that asymmetric false positives also drive the global FPR close to 1, but they lead to worse outcomes than the symmetric case, especially for OLMo. Because the verifier now induces a one-sided training signal, this suggests that the asymmetric false-positive interval systematically pushes the model toward over- or underestimation.

This comparison also indicates that one cannot solely evaluate verifiers based on simple metrics at the start of training. In our setting, an asymmetric interval rewards a strictly smaller region and therefore has a lower initial FPR than a symmetric one, yet it still produces worse training outcomes. This highlights a key limitation of evaluating verifiers based on their average error rate: the effect of verifier errors on training depends not just on their overall rate, but on how they distort the learning dynamics.

\subsection{Widespread False Negatives}\label{subsec:language_fn}

As shown in \cref{subsec:main_result}, systematic false negatives can delay training. Here, we show that even a very high FNR may still delay training rather than cause collapse. As an extreme example of systematic false negatives, we consider a language-based verifier that assigns a false negative whenever the generated output is in English. In this setting, the model can receive a positive reward only if its answer is both correct and written in a non-English language. We use \texttt{langdetect} library~\citep{langdetect} to identify language of each output.

\paragraph{Results} \Cref{fig:language_fn} shows the results of this experiment. Although the global FNR begins close to 1, it declines quickly once the model learns to produce non-English answers, allowing it to improve performance. As a result, the training curve shows a delayed pattern similar to the format-based false negatives in our main results, though the delay is more pronounced here because the model must first learn to avoid English before it can begin improving task performance. Under this constraint, Qwen learns to answer in Chinese, whereas OLMo instead exploits \texttt{langdetect} by inserting repetitive English words into its response, both of which effectively bypass the language-based false negative.

Importantly, false negatives do not always lead to such behaviors. For example, if a verifier consistently assigns negative rewards to all queries from a domain, training on that domain can do no more than plateau, since the model receives no useful learning signal there.

\subsection{Mitigation Strategy: Alternation with Ground-Truth Verifier}
\label{subsec:oracle_alternation}

\begin{wrapfigure}{r}{0.6\textwidth}
    \centering
    \begin{subfigure}{\linewidth}
        \centering
        \includegraphics[width=0.65\linewidth]{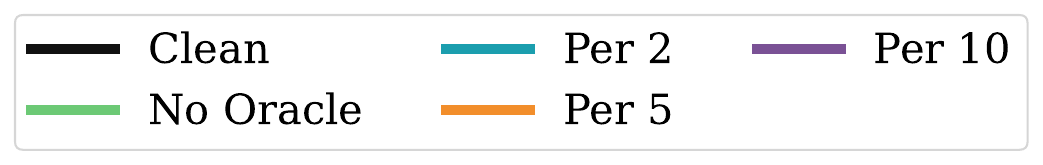}
        \includegraphics[width=\linewidth]{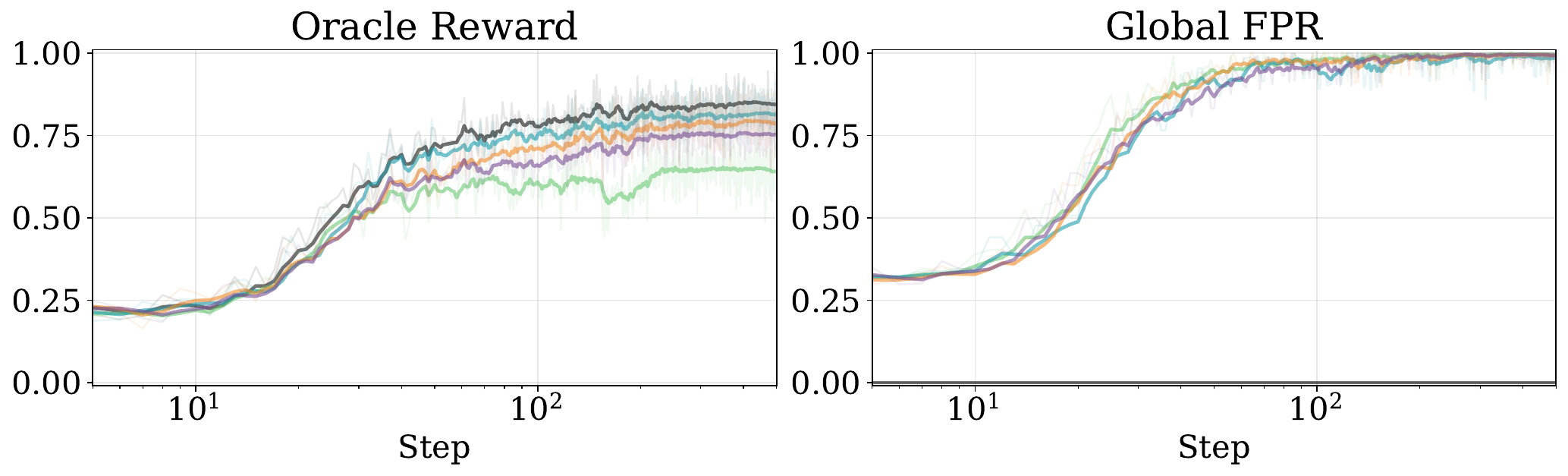}
        \caption{Relative error-based FP}
        \label{fig:alternation_oracle_relative_error}
    \end{subfigure}

    \begin{subfigure}{\linewidth}
        \centering
        \includegraphics[width=\linewidth]{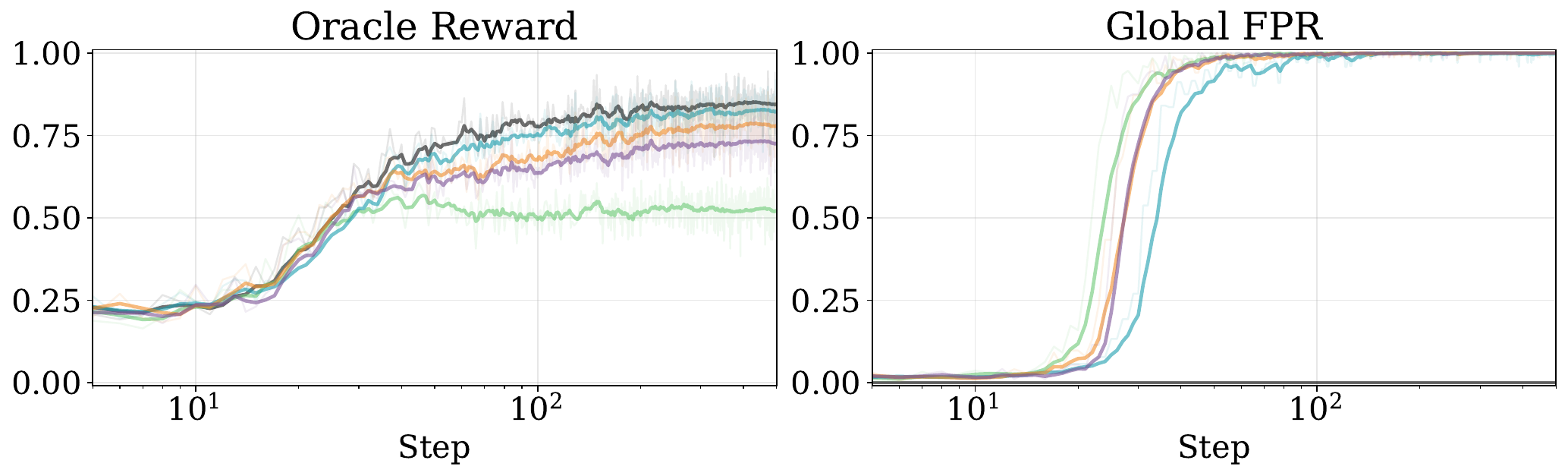}
        \caption{\texttt{"Certainly"}-triggered FP}
        \label{fig:alternation_oracle_certainly}
    \end{subfigure}

    \begin{subfigure}{\linewidth}
        \centering
        \includegraphics[width=\linewidth]{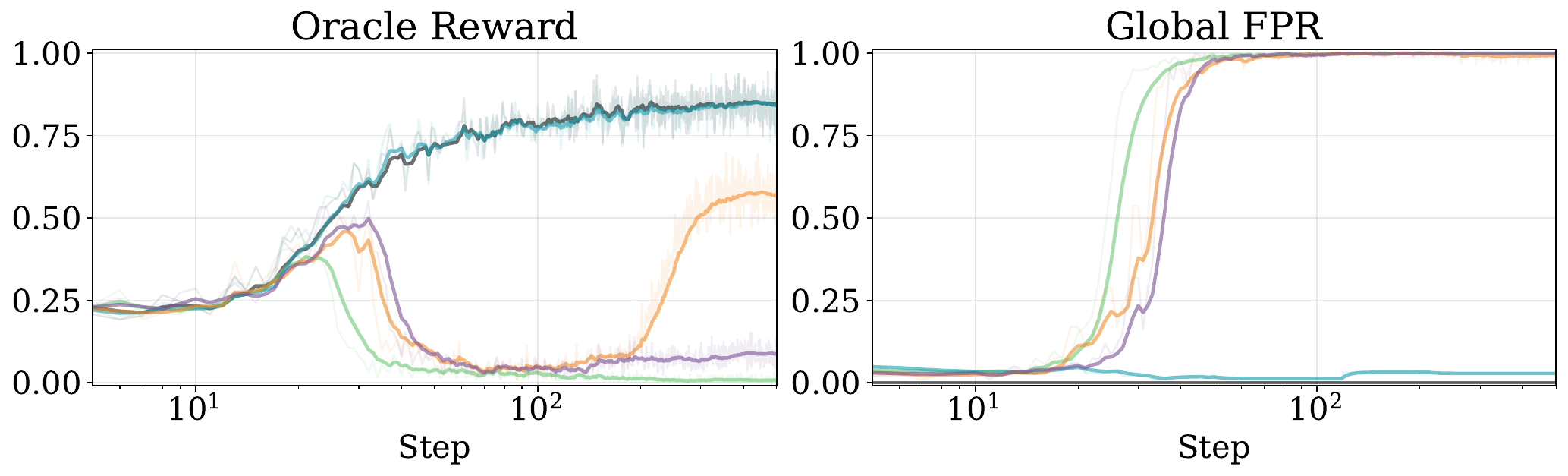}
        \caption{\texttt{"python"}-triggered FP}
        \label{fig:alternation_oracle_python}
    \end{subfigure}
    \vspace{-1.5em}
    \caption{
        Alternation of oracle and noisy verifier.
        FPR is measured only for the steps with the noisy verifier.
    }
    \label{fig:alternation}
    \vspace{-1.5em}
\end{wrapfigure}

Training longer cannot reliably overcome systematic errors, as the model may plateau or even collapse. But can sparse access to a ground-truth verifier stabilize training or mitigate collapse? To answer this, we study a simple strategy that alternates between an imperfect verifier and the ground-truth verifier during training.

\paragraph{Results}
Results are shown in \cref{fig:alternation}. There is a clear difference between plateauing training dynamics and collapsing ones. As shown in \cref{fig:alternation_oracle_relative_error,fig:alternation_oracle_certainly}, the alternation mostly mitigates the effect of verification errors and ensures the reward becomes more akin to a delayed curve. This occurs even when the oracle verifier is only used once every 10 steps, which is a relatively small fraction of training. In contrast, the alternation has a much weaker effect in the collapsing case (\cref{fig:alternation_oracle_python}). With sparse oracle access, training still collapses, albeit slightly more slowly. With more frequent access (once every 2 steps), collapse is prevented.

However, in most cases, the alternation does not eliminate the trigger behavior itself, and FPR still increases to $1$. Instead, the model learns to make the trigger pattern coexist with correct reasoning. Since this causes the imperfect verifier to give positive reward for all outputs, the training essentially optimizes the oracle reward when the oracle verifier is used, which prevents collapse and leads to a delayed curve.

\section{Discussion}
\label{sec:discussion}
\paragraph{Implications for RLVR and verifier design}
As RLVR is increasingly applied in real-world settings, our findings highlight the importance of understanding and mitigating verification errors. Even on a relatively simple task, different error patterns can lead to qualitatively different training dynamics, including delayed learning, plateauing, and collapse. These dynamics are also difficult to predict from aggregate measures of verifier quality, such as overall error rate or Youden's $J$ statistic \citep{rad2026rate}, which do not capture the systematic nature of the errors. It is therefore crucial to develop more nuanced diagnostics that can identify and characterize systematic error patterns, as well as mitigation strategies.

\paragraph{Limitations and future work}
Our study is conducted in a controlled setting with manually specified error patterns, which is useful for isolating the effect of systematic verification errors and measuring oracle reward. However, practical RLVR settings are more complex: real verifiers may exhibit multiple interacting failure modes and depend on the input as well as the output.
An important next step is therefore to test these phenomena in domains such as code generation, formal reasoning, and rubric-based rewards, and to develop diagnostics and mitigation strategies that remain effective without full oracle access.

\section{Conclusion}
\label{sec:conclusion}
In this work, we analyze the impact of \textit{systematic} verification errors in RLVR, where the verifier consistently assigns the wrong reward signal to outputs with a certain property.
Unlike the random noise studied in prior work, we find that systematic errors can have markedly different effects depending on their type: training may be delayed, plateau, or collapse.
Through controlled experiments and analysis, we show what types of verification errors lead to what outcomes, and what determines the fate of training under systematic errors.
We hope this work motivates future research to account for these differences and to characterize verifier error patterns in ways that go beyond aggregate error rates.

\section*{Acknowledgments}
This work has been done as part of the SERI grant SAFEAI (Certified Safe, Fair and Robust Artificial Intelligence, contract no. MB22.00088). Views and opinions expressed are however those of the authors only and do not necessarily reflect those of the European Union or European Commission. Neither the European Union nor the European Commission can be held responsible for them. The work has received funding from the Swiss State Secretariat for Education, Research and Innovation (SERI) (SERI-funded ERC Consolidator Grant). This work was supported as part of the Swiss AI Initiative by a grant from the Swiss National Supercomputing Centre (CSCS) under project ID a155 and a158 on Alps.
Florian Dorner is grateful for financial support from the Max Planck ETH Center for Learning Systems (CLS).

\section*{Reproducibility Statement}
We provide detailed descriptions of our experimental setup, including the task, model, training algorithm, and noise patterns in \cref{app:experimental_details}. The code for our experiments is publicly available at \url{https://github.com/eth-sri/llm-verifier-noise}. It includes comprehensive instructions for reproducing our results, along with the exact hyperparameters used for training and the implementation of the noise patterns.

\bibliography{colm2026_conference}
\bibliographystyle{unsrtnat}

\message{^^JLASTREFERENCESPAGE \thepage^^J}

\ifincludeappendixx
	\newpage
	\appendix
	\onecolumn
	
\section{More Details on Experimental Setting}\label{app:experimental_details}

\subsection{Additional Details on Error Patterns}
\label{appsubsec:error_design}

\paragraph{Format-Based FN}
Here, a correct answer is incorrectly rejected when it appears in a particular output format. This models rule-based verifiers that fail on valid but unexpected answer representations. Concretely, we use the most common formatting pattern observed under clean-verifier training, namely the LaTeX-style token \texttt{\textbackslash[}. We consider two variants: one in which an FN is triggered when the output contains this pattern, and another in which an FN is triggered when the output does not contain it. For example, a model may output \texttt{"The answer is \textbackslash[ \textbackslash boxed\{X\} \textbackslash]"}; in the first variant of the error design, this output receives reward 0 even if \texttt{X} is correct.

\paragraph{Relative Error-Based FP}
In this setting, an incorrect answer can still receive a positive reward if its relative error from the ground-truth answer, \ie $\lvert \text{prediction} - \text{ground-truth} \rvert / \lvert \text{ground-truth} \rvert$, falls below a fixed threshold. This models verifiers that accept approximate answers rather than requiring exact equality. Such behavior can arise when outputs are evaluated up to a tolerance, for example, because of limited measurement precision, sensor noise, or evaluation pipelines that treat sufficiently close values as correct. We consider thresholds of $0.1$ and $1$.

\paragraph{Word-Based FP}
Here, the verifier assigns positive reward whenever the model output contains a predefined keyword.
This setup is motivated by recent findings that LLM-based verifiers can be manipulated by the presence of a single token~\citep{zhao2025one}.

\paragraph{Language-Based FN}
In this setting, the verifier assigns zero reward whenever the model output is classified as English based on \texttt{langdetect}~\citep{langdetect} library.
Compared to other error patterns that are more strongly inspired by real-world scenarios, this provides an extreme example where the verifier fails to recognize correct answers in almost all generations, resulting in almost no learning signal at the beginning.

\begin{wraptable}{r}{0.4\linewidth}
\vspace{-2em}
\centering
\caption{Training Configuration}
\label{apptab:training_hyperparameters}
\resizebox{\linewidth}{!}{
    \begin{tabular}{ll}
        \toprule
        \textbf{Dataset} & \\
        \midrule
        ~~Dataset size & 50,000 \\
        ~~min\_terms & 3 \\
        ~~max\_terms & 6 \\
        ~~min\_digits & 3 \\
        ~~max\_digits & 6 \\
        ~~min\_decimal\_places & 3 \\
        ~~max\_decimal\_places & 6 \\
        ~~allow\_negation & False \\
        \midrule
        \textbf{Training Arguments} & \\
        \midrule
        ~~Effective batch size & 256 \\
        ~~Learning rate & $5 \times 10^{-6}$ \\
        ~~Rollout per query & 4 \\
        ~~Max completion length & 2048 \\
        ~~Max step & 500 \\
        ~~Sampling temperature & 1.0 \\
        ~~Algorithm & AdaFactor \\
        ~~Epoch & 1 \\
        ~~Scheduler & Cosine \\
        ~~Warmup steps & 20 \\
        \bottomrule
    \end{tabular}
}

\vspace{-2em}
\end{wraptable}

\subsection{Training Hyperparameters}
\label{appsubsec:training_hyperparameters}

As briefly noted in~\Cref{subsec:setting}, we use the Reasoning Gym dataset~\citep{stojanovski2025reasoning}.
\Cref{apptab:training_hyperparameters} lists the training hyperparameters.
Because we can generate unlimited data with Reasoning Gym, we set a large dataset size (50,000) so that the model receives new samples at each step.
Since training requires 64 unique queries per batch (256 batch size divided by 4 rollouts each), the total number of unique queries seen during training is 32,000.
Through preliminary tuning, we set the difficulty so that both models start at roughly 0.2 reward and reach 0.9 under a clean verifier.

\paragraph{Training Framework}
We use Hugging Face TRL~\citep{vonwerra2020trl} and vLLM for sampling~\citep{vllm}.

\paragraph{Prompt}

\begin{figure*}[t]
\centering

\begin{minipage}{\textwidth}
\begin{promptbox}[title=Example Query,colback=black!10!white, colframe=black!90!white]
\small
\begin{wrapverb}
State the final answer to the following arithmetic problem: 481.869 + 519.413 + 776.7711 - 734.133 =
Please reason step by step, and put your final answer within \boxed{}.
\end{wrapverb}
\end{promptbox}
\end{minipage}

\caption{
	An example query used for training.
}
\label{appfig:query_example}
\end{figure*}

We prompt the model to wrap its answer with \texttt{\textbackslash boxed\{\}}, and compare the generated answer with the correct answer by extracting the content within the box.
An example prompt is shown in~\Cref{appfig:query_example}.

\subsection{Trigram selection}
\label{appsubsec:trigram_setting}
Since the experiment in~\Cref{fig:token_dot} is costly (one training run per point), we selected trigrams as follows.
We first queried the base model with 9,216 samples from the training set (16 rollouts for each of 576 questions) to estimate trigram frequency and conditional advantage.
We then kept trigrams with frequency between 5\% and 15\%, yielding 509 candidates.
After removing trigrams containing non-alphabetic tokens (\eg \texttt{is:\textbackslash n\textbackslash n \textbackslash}), which are harder to associate with the behaviors of interest, 148 remained.

Among these, 29 trigrams had positive conditional advantage; we trained one model for each, using it as a systematic FP trigger.
Of the 119 trigrams with negative conditional advantage, we sampled 22 and trained one model per sampled trigram.

For these runs, we trained for 300 steps instead of 500 and imposed a 4-hour time limit, since some collapsing runs produced very long completions and led to long training times.
As a result, 18 runs ended before reaching 300 steps, with the shortest stopping at 186 steps.

\newpage
\section{More Results}\label{appsec:more_results}

\subsection{Length-based False Positives}
\label{appsubsec:more_error_type}

\begin{figure}[t]
\centering

\begin{minipage}[t]{0.49\textwidth}
    \centering
    \vspace{0pt}
    \begin{subfigure}{\linewidth}
        \centering
        \includegraphics[width=0.6\linewidth]{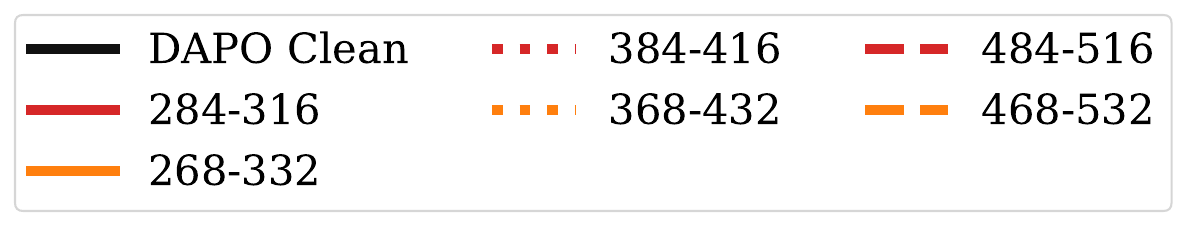}
        \includegraphics[width=\linewidth]{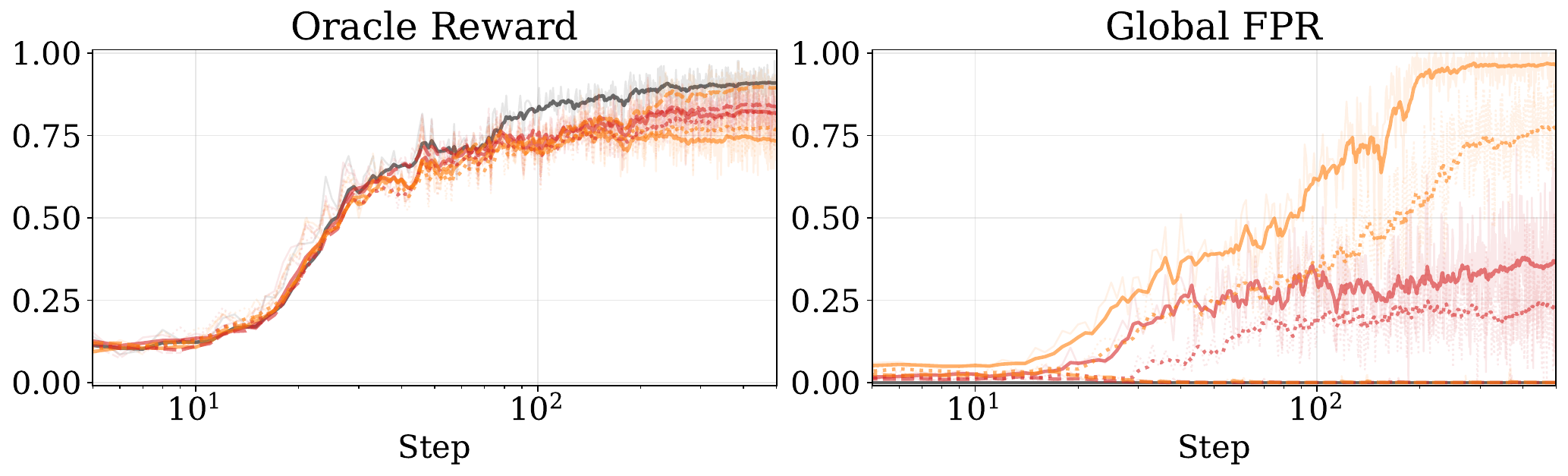}
        \caption{OLMo}
        \label{appsubfig:length_fp_olmo}
    \end{subfigure}

    \vspace{0.2em}

    \begin{subfigure}{\linewidth}
        \centering
        \includegraphics[width=0.6\linewidth]{figures/length_fp/olmo_length_fp_legend.pdf}
        \includegraphics[width=\linewidth]{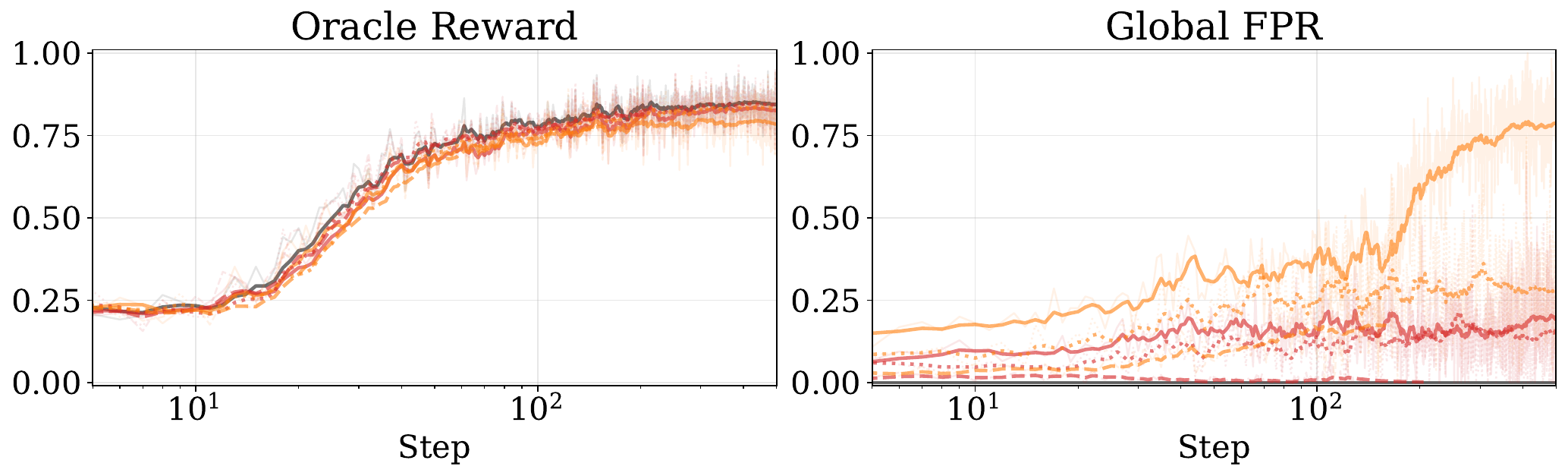}
        \caption{Qwen}
        \label{appsubfig:length_fp_qwen}
    \end{subfigure}

    \vspace{0.4em}

    \begin{minipage}{0.98\linewidth}
        \caption{
            Results on length-based FP.
            The legend indicates the interval of the output completion within which the verifier gives a FP.
        }
        \label{appfig:length_fp}
    \end{minipage}
\end{minipage}
\hfill
\begin{minipage}[t]{0.49\textwidth}
    \centering
    \vspace{0pt}
    \begin{subfigure}{\linewidth}
        \centering
        \includegraphics[width=0.9\linewidth]{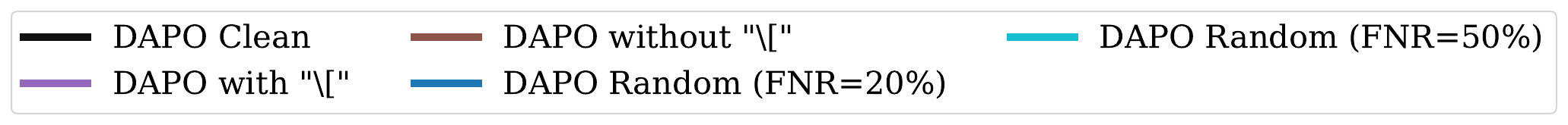}\par
        \vspace{0.8em}
        \includegraphics[width=\linewidth]{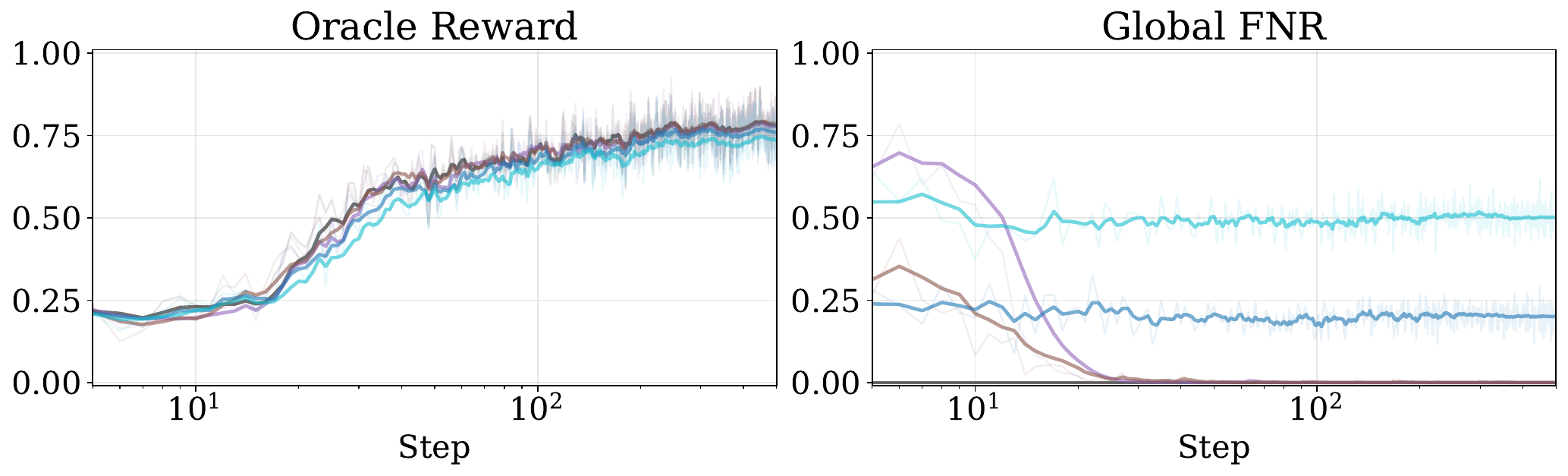}
        \caption{FNR}
        \label{fig:instruct_qwen_fnr}
    \end{subfigure}

    \vspace{0.4em}

    \begin{subfigure}{\linewidth}
        \centering
        \includegraphics[width=0.9\linewidth]{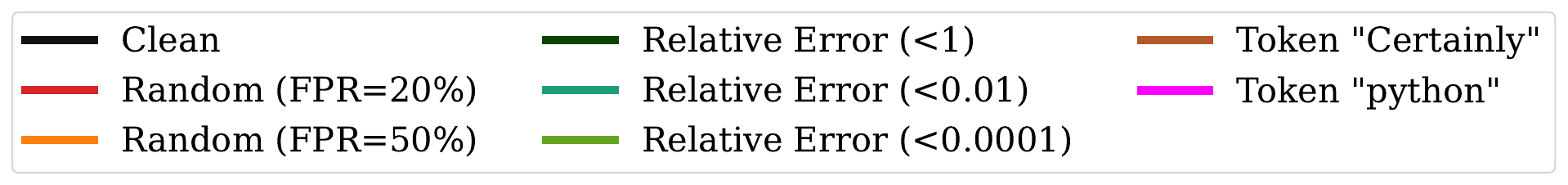}
        \includegraphics[width=\linewidth]{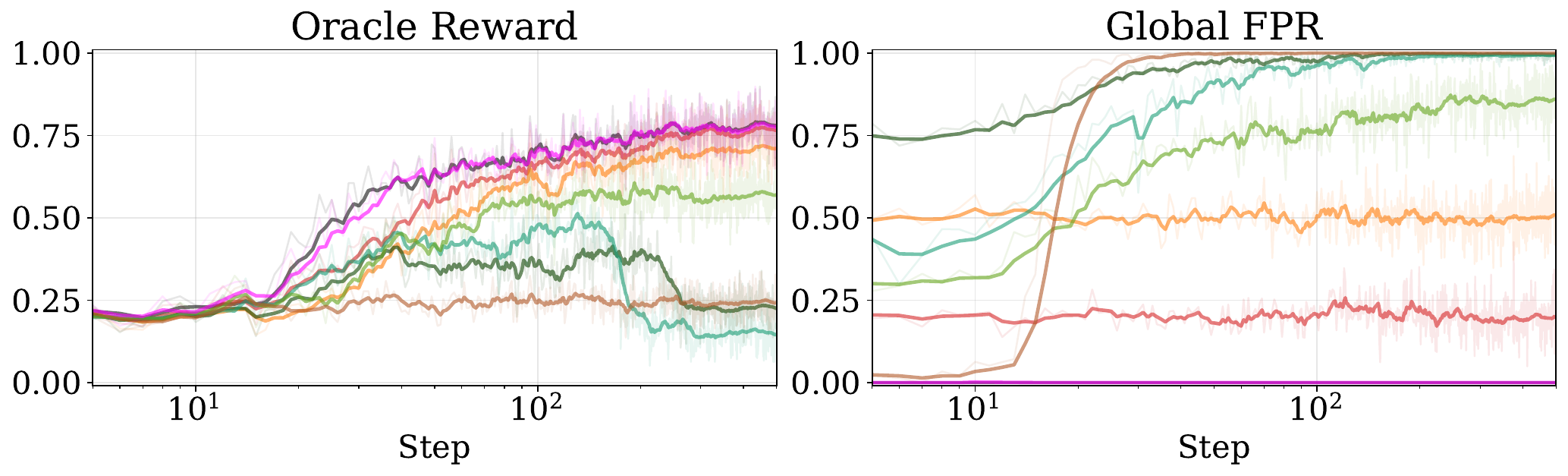}
        \caption{FPR}
        \label{fig:instruct_qwen_fpr}
    \end{subfigure}

    \vspace{0.4em}

    \begin{minipage}{0.98\linewidth}
        \caption{
            Results on Qwen2.5-1.5B-Instruct.
        }
        \label{fig:instruct_qwen}
    \end{minipage}
\end{minipage}

\end{figure}

Some systematic FPs are inherently more difficult to learn to exploit than others.
In such cases, the FPR can remain relatively low throughout training, and the effect of the noise can be more similar to a delaying one than a plateauing one.
Length-based FP, where the verifier introduces FPs if the output falls within a certain length interval, is one such example.
In~\Cref{appfig:length_fp}, we present the results with length-based FP.

Most settings show a delayed training dynamic, with the FPR staying relatively low, below 0.3, throughout training. Some settings, however, exhibit a more plateau-like behavior, with the FPR rising to around 0.5. This tends to occur for larger intervals, which naturally makes the hack easier for the model to exploit.

Additionally, some lengths are easier to exploit than others. An FP interval of around 300 consistently produces higher FPR and, therefore, more severe plateauing than the other intervals. This is likely related to the fact that the average output length under clean-verifier training converges to around 300, making it easier for the model to hack the verifier when the FP interval is near that length.

\subsection{Results on Instruction-tuned Models}
\label{appsubsec:instruct}

In the main paper, we present results for the base models. Here, we report the results obtained when starting from an instruction-tuned model, Qwen2.5-1.5B-Instruct.

For random noise and systematic FN, we observe a delayed training trend that is consistent with the base model results.

In contrast, systematic FP leads to qualitatively different behavior. With \texttt{python}-triggered FP, the training curve is similar to that of the clean verifier, consistent with a 0\% FPR: the model never generated the token \texttt{"python"} during training. With \texttt{Certainly}-triggered FP, the model reaches a much lower plateau than the base model. A likely explanation is that the instruction-tuned model can produce many correct answers without beginning with \texttt{"Certainly"}, so this token is not strongly associated with correctness, and rewarding it offers little benefit, though it still does not cause collapse.

For relative-error-based FP, the results are mixed. At larger thresholds, training shows a collapsing trend. We further reduce the threshold to 0.0001 and find that the curve plateaus throughout training only at this value. At all other thresholds, training still collapses, likely because the instruction-tuned model is already capable of generating nearly correct answers, so rewarding such outputs does not meaningfully encourage further improvement.
\subsection{Test Performance}
\label{appsubsec:test_performance}

\begin{table}
\centering
\caption{
    Pass@k results under different noise settings.
}
\vspace{-3mm}
\label{apptab:pass_at_k}
\begin{adjustbox}{width=0.6\linewidth}
\begin{tabular}{llllll}
\toprule
Model & Error Category & Setting & Pass@1 & Pass@5 & Pass@16 \\
\midrule
Qwen & Base Model & temperature=1 & 20.6 & 60.6 & 86.2 \\
 &  & temperature=0.6 & 53.3 & 85.0 & 93.8 \\
 \cmidrule{2-6}
 & Clean & - & 85.6 & 89.3 & 91.0 \\
 \cmidrule{2-6}
 & Random Noise & FPR=20\% & 84.0 & 88.1 & 89.6 \\
 &  & FPR=50\% & 82.2 & 87.4 & 89.2 \\
 &  & FNR=20\% & 85.3 & 89.1 & 90.4 \\
 &  & FNR=50\% & 84.1 & 88.3 & 89.4 \\
 \cmidrule{2-6}
 & Systematic FP & Relative Error < 1 & 66.9 & 81.1 & 86.4 \\
 &  & Relative Error < 0.1 & 69.2 & 82.7 & 87.4 \\
 &  & Token "python" & 0.6 & 2.2 & 4.0 \\
 &  & Token "Certainly" & 64.3 & 84.0 & 90.0 \\
 \cmidrule{2-6}
 & Systematic FN & FN if "\textbackslash{}[" & 85.2 & 88.0 & 88.8 \\
 &  & FN if no "\textbackslash{}[" & 85.1 & 89.0 & 90.2 \\
 \midrule
OLMo & Base Model & temperature=1 & 16.2 & 51.1 & 75.8 \\
 &  & temperature=0.6 & 28.3 & 65.0 & 81.0 \\
 \cmidrule{2-6}
 & Clean & - & 93.5 & 95.4 & 96.2 \\
 \cmidrule{2-6}
 & Random Noise & FPR=20\% & 89.7 & 92.6 & 93.6 \\
 &  & FPR=50\% & 87.4 & 91.4 & 92.8 \\
 &  & FNR=20\% & 89.2 & 92.3 & 93.4 \\
 &  & FNR=50\% & 80.0 & 85.1 & 87.0 \\
 \cmidrule{2-6}
 & Systematic FP & Relative Error < 1.0 & 50.8 & 62.3 & 68.2 \\
 &  & Relative Error < 0.1 & 54.3 & 65.1 & 70.4 \\
 &  & Token "python" & 2.0 & 8.4 & 21.2 \\
 &  & Token "Certainly" & 54.3 & 74.0 & 82.6 \\
 \cmidrule{2-6}
 & Systematic FN & FN if "\textbackslash{}[" & 93.0 & 94.5 & 95.2 \\
 &  & FN if no "\textbackslash{}[" & 84.6 & 88.0 & 90.0 \\
\bottomrule
\end{tabular}
\end{adjustbox}

\end{table}

We present benchmarking results for the trained models under a range of error patterns. For this evaluation, we use the same decimal chain sum task and difficulty configuration as in the main text. We generate 500 questions with a different random seed, query each trained model with 16 rollouts per question, and compute Pass@k for $k=1,5,16$. The results are shown in \cref{apptab:pass_at_k}.
We note that training uses the default sampling temperature of 1. As a result, the initial reward is similar to the base model's Pass@1 at temperature 1.

\subsection{Qualitative Comparison Between Trigrams}
\label{appsubsec:trigram_qualitative}
\begin{figure*}[t]
\centering

\begin{subfigure}[t]{0.48\textwidth}
	\centering
	\input{prompts/trigram/add_the_first.tex}
	\caption{CA=0.21, Freq=5.1\%, R=0.72}
\end{subfigure}
\hfill
\begin{subfigure}[t]{0.48\textwidth}
	\centering
	\input{prompts/trigram/to_the_result.tex}
	\caption{CA=0.27, Freq=12.1\%, R=0.68}
\end{subfigure}

\vspace{0.5em}

\begin{subfigure}[t]{0.48\textwidth}
	\centering
	\input{prompts/trigram/let_me_verify.tex}
    \caption{CA=$-$0.14, Freq=5.9\%, R=0.89}
\end{subfigure}
\hfill
\begin{subfigure}[t]{0.48\textwidth}
	\centering
	\input{prompts/trigram/reason_step_by.tex}
    \caption{CA=$-$0.13, Freq=12.8\%, R=0.02}
\end{subfigure}
\caption{
    Examples for learned behaviors in the trigram-based error patterns.
    For a question \texttt{481.869 + 519.413 + 776.7711 - 734.133}, (ground truth: 1043.9201), we provide a representative output example for positive / negative conditional advantage (defined in~\Cref{subsec:word_based_fp}), and low / high frequency.
    For each caption, we also provide the conditional advantage (CA), frequency (Freq), and the converged reward (R) of the when the trigram is used as the FP trigger.
}
\label{appfig:trigram_compare}
\end{figure*}

We qualitatively compare the behaviors learned under different trigram-based FP error patterns, which give rise to distinct training dynamics, in \cref{appfig:trigram_compare}.

When the trigger is a trigram with negative conditional advantage, e.g., \texttt{"reason step by"}, the model begins to repeat the prompt instead of producing the final answer, leading to substantial performance degradation from the oracle perspective.

In contrast, when the trigger is a frequent trigram with positive conditional advantage, e.g., \texttt{"to the result"}, the model tends to say \texttt{"add X to the result"} whenever it encounters an addition step. This behavior still produces the correct answer and therefore leads to a plateauing reward.

When the trigger is less frequent and has a negative conditional advantage, as in \texttt{"Let me verify"}, the model produces correct answers without generating the specified trigram, resulting in a reward level similar to that of the clean verifier.

Finally, a less frequent trigram with positive conditional advantage, such as \texttt{"add the first"}, can still be amplified because it aligns with the oracle reward and is learned alongside correct behavior, which again leads to a plateau.

\subsection{Results with Different Algorithms}
\label{appsubsec:different_algorithms}

\begin{figure}[t]
\centering
\setlength{\tabcolsep}{3pt}
\resizebox{0.95\linewidth}{!}{
\begin{tabular}{cc}
\textbf{OLMo3-7B} & \textbf{Qwen3-1.7B-Base} \\
\midrule

\multicolumn{2}{c}{\textbf{DAPO}} \\
\multicolumn{2}{c}{\textit{FNR}} \\
\multicolumn{2}{c}{\includegraphics[width=0.7\textwidth]{figures/main_result/dapo/olmo_fnr_legend.pdf}} \\
\includegraphics[width=0.48\textwidth]{figures/main_result/dapo/olmo_fnr_without_legend.pdf} &
\includegraphics[width=0.48\textwidth]{figures/main_result/dapo/qwen_fnr_without_legend.pdf} \\

\multicolumn{2}{c}{\textit{FPR}} \\
\multicolumn{2}{c}{\includegraphics[width=0.7\textwidth]{figures/main_result/dapo/olmo_fpr_legend.pdf}} \\
\includegraphics[width=0.48\textwidth]{figures/main_result/dapo/olmo_fpr_without_legend.pdf} &
\includegraphics[width=0.48\textwidth]{figures/main_result/dapo/qwen_fpr_without_legend.pdf} \\

\midrule

\multicolumn{2}{c}{\textbf{Dr. GRPO}} \\
\multicolumn{2}{c}{\textit{FNR}} \\
\multicolumn{2}{c}{\includegraphics[width=0.7\textwidth]{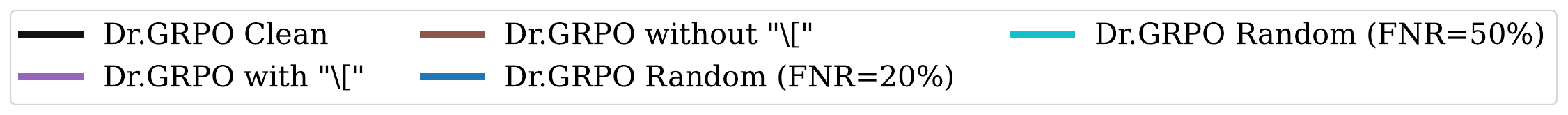}} \\
\includegraphics[width=0.48\textwidth]{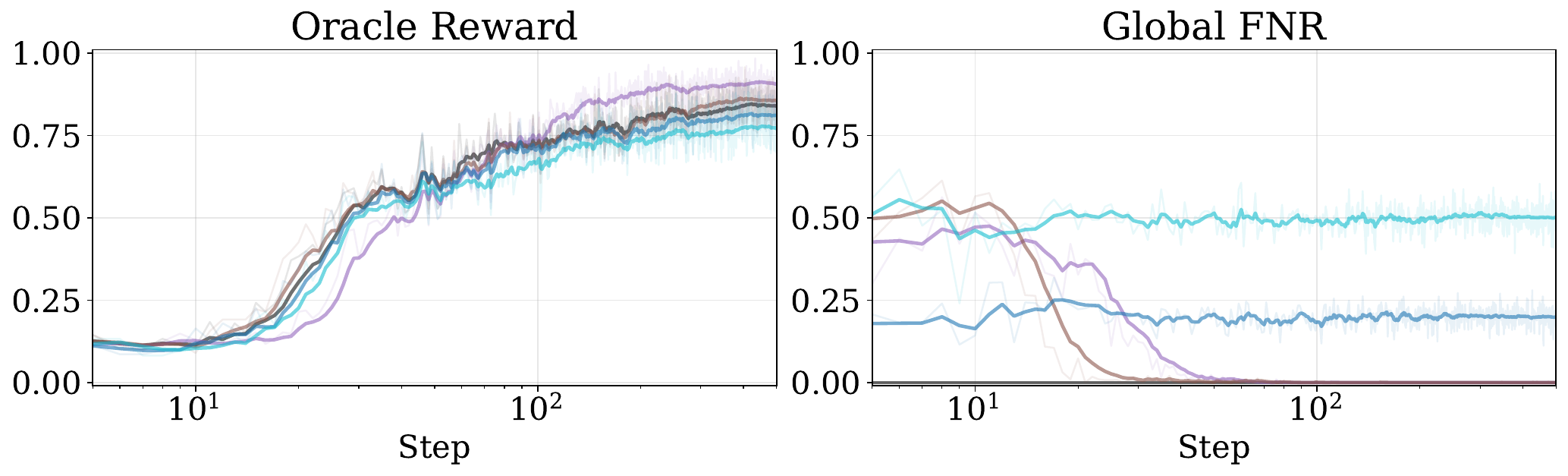} &
\includegraphics[width=0.48\textwidth]{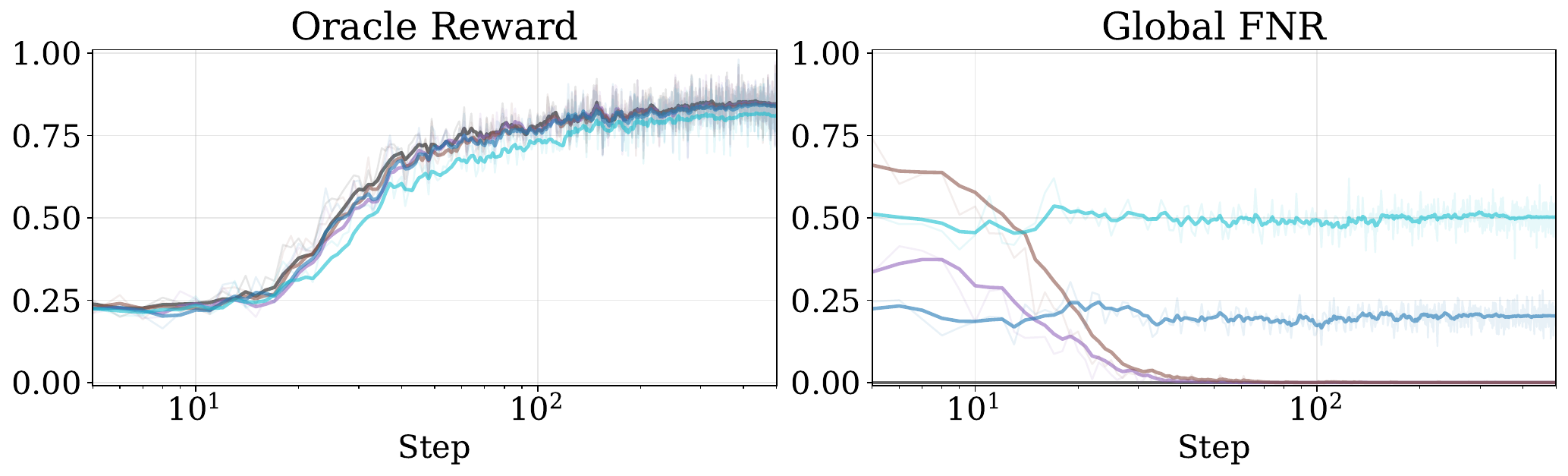} \\

\multicolumn{2}{c}{\textit{FPR}} \\
\multicolumn{2}{c}{\includegraphics[width=0.7\textwidth]{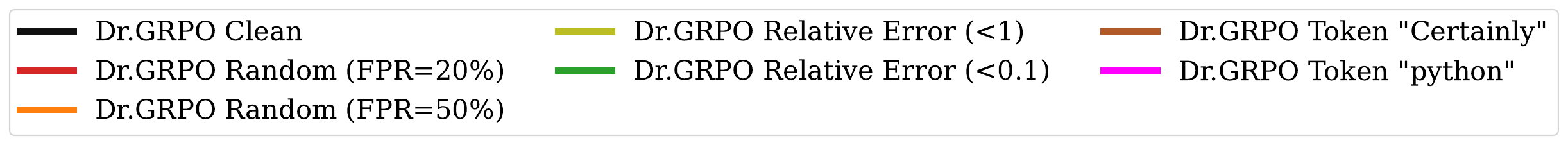}} \\
\includegraphics[width=0.48\textwidth]{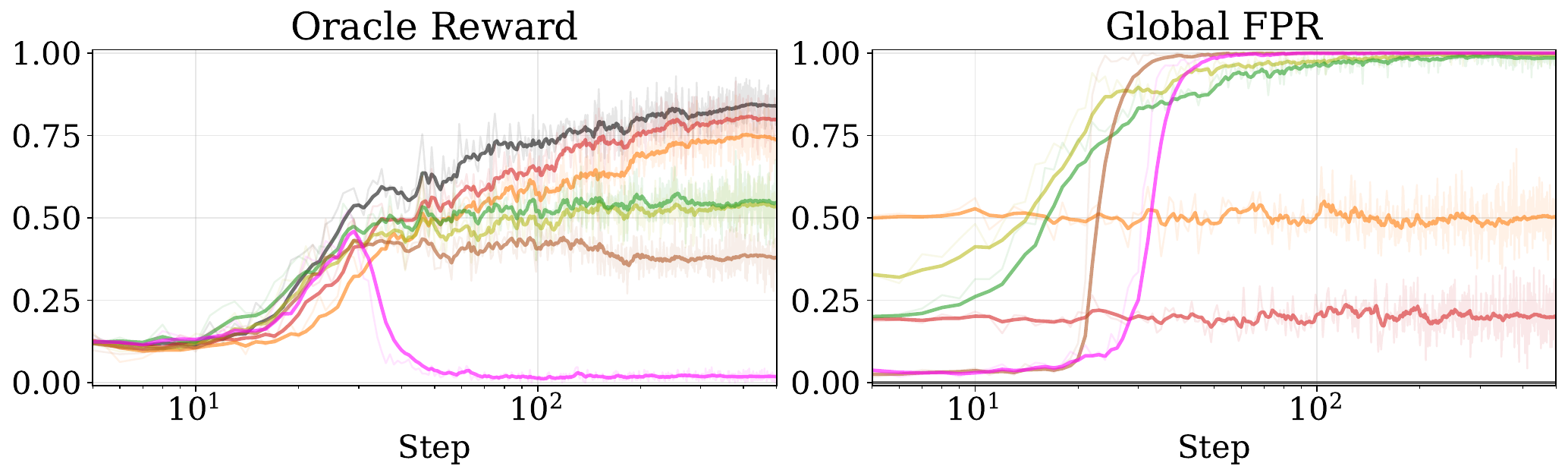} &
\includegraphics[width=0.48\textwidth]{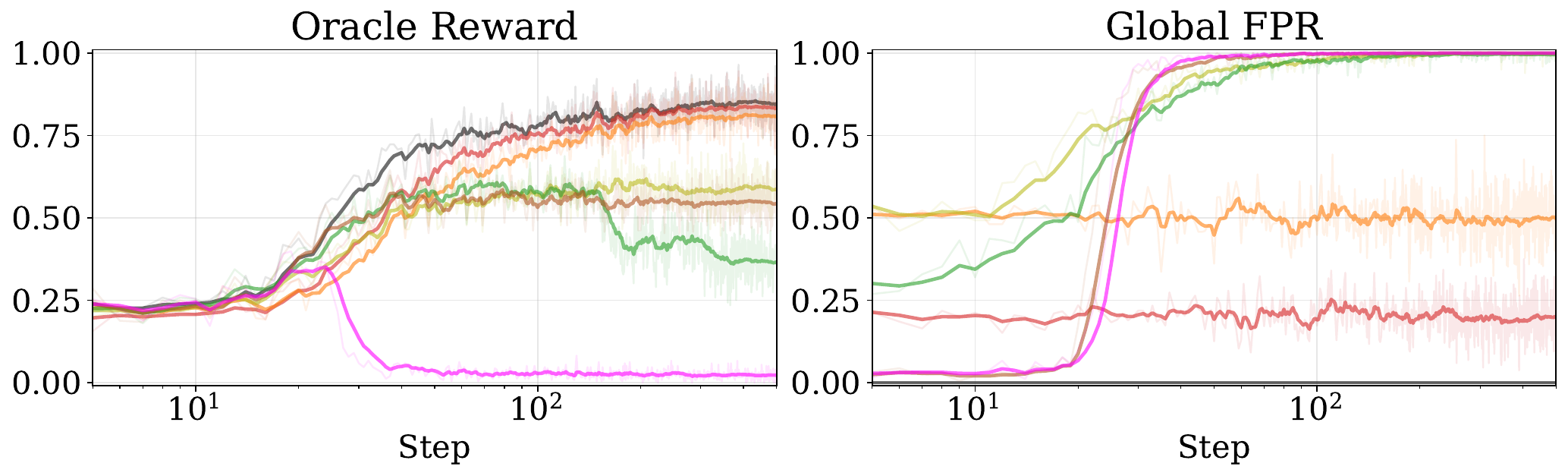} \\

\midrule

\multicolumn{2}{c}{\textbf{SAPO}} \\
\multicolumn{2}{c}{\textit{FNR}} \\
\multicolumn{2}{c}{\includegraphics[width=0.7\textwidth]{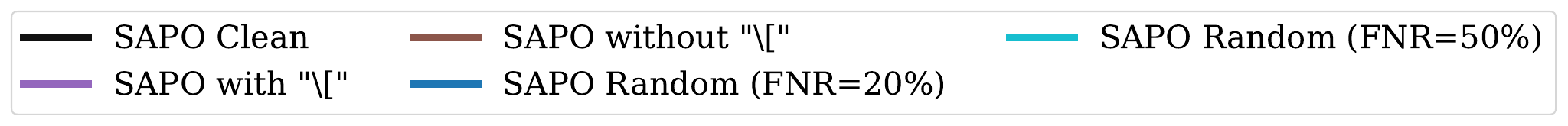}} \\
\includegraphics[width=0.48\textwidth]{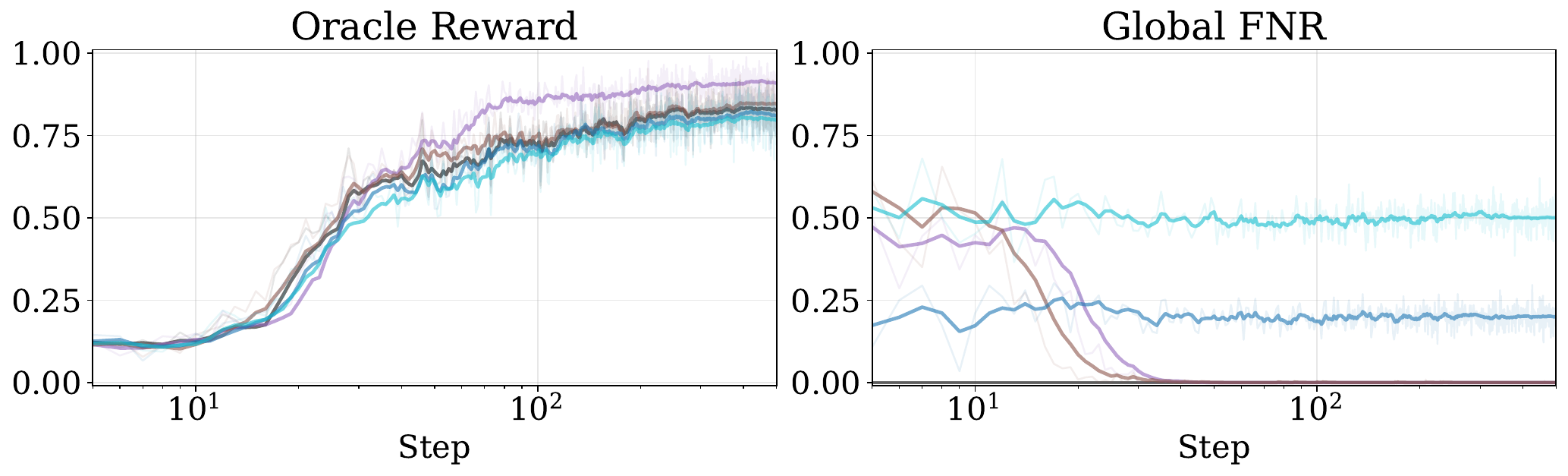} &
\includegraphics[width=0.48\textwidth]{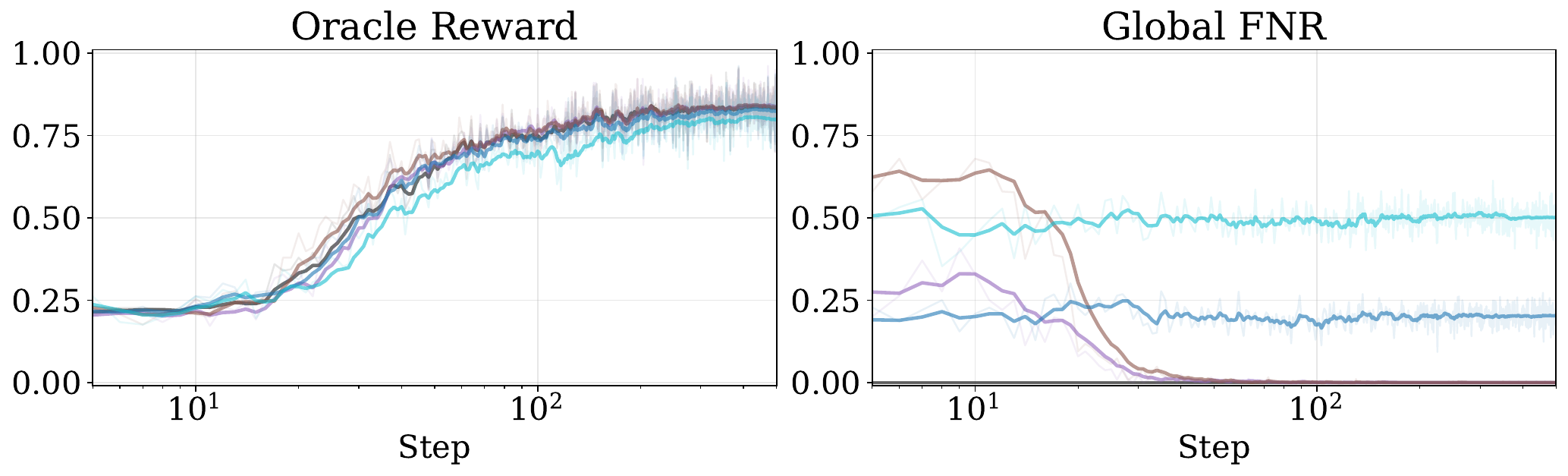} \\

\multicolumn{2}{c}{\textit{FPR}} \\
\multicolumn{2}{c}{\includegraphics[width=0.7\textwidth]{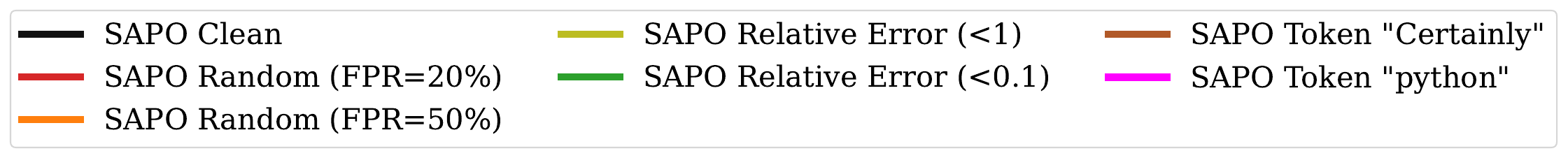}} \\
\includegraphics[width=0.48\textwidth]{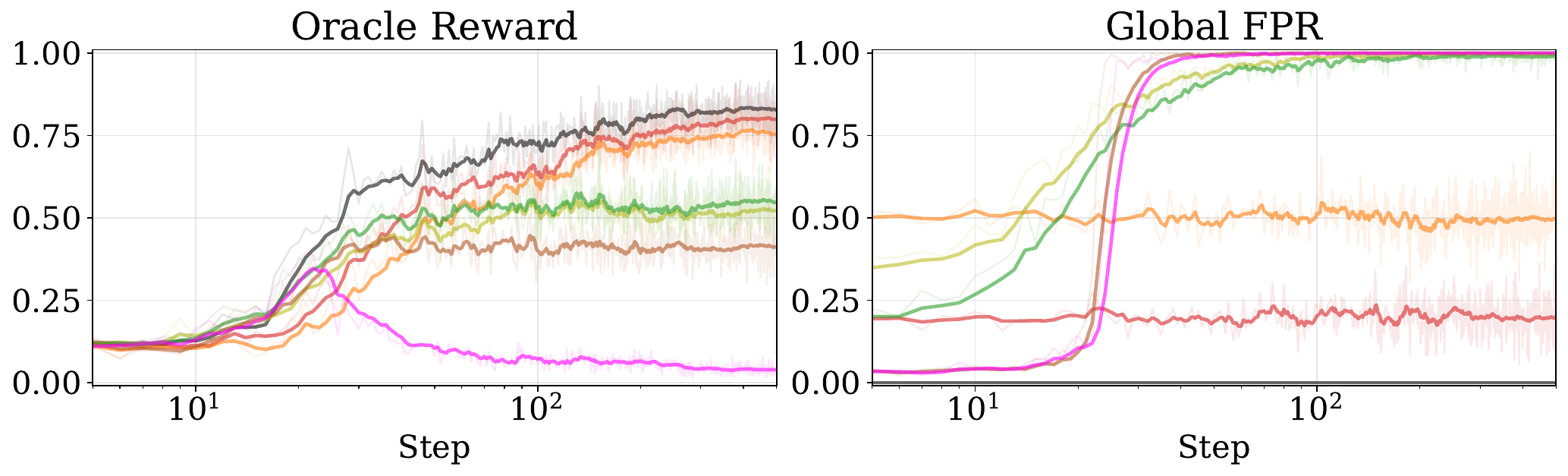} &
\includegraphics[width=0.48\textwidth]{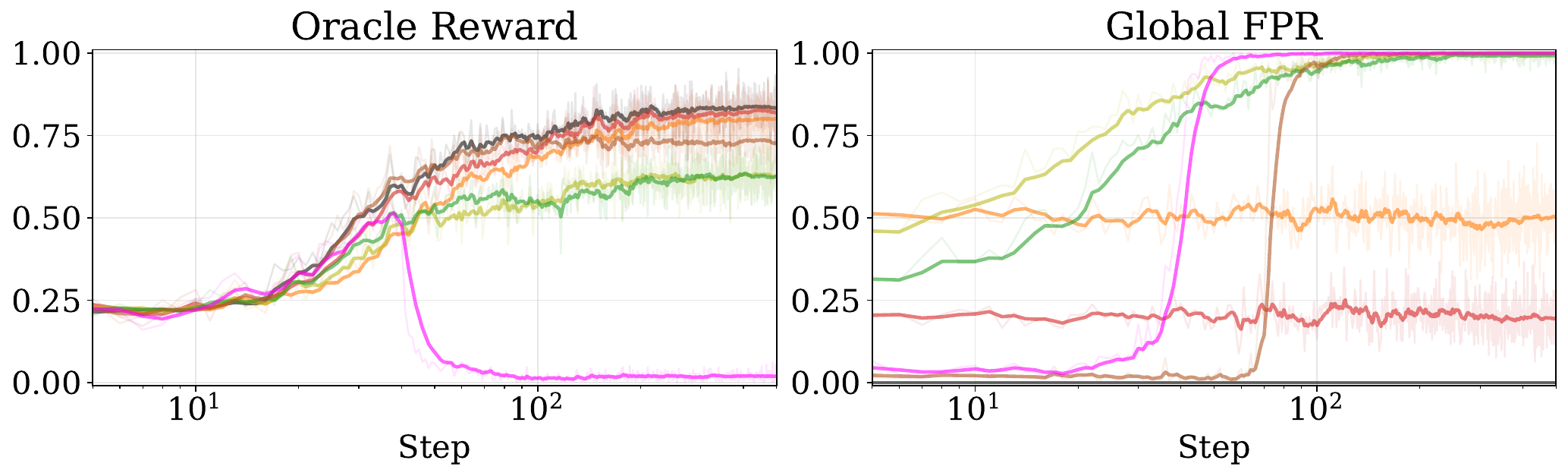} \\

\end{tabular}
}
\caption{
    Results with DAPO, Dr. GRPO, and SAPO.
}
\label{appfig:all_algorithms}
\renewcommand{\arraystretch}{1.0}
\end{figure}

In~\Cref{appfig:all_algorithms}, we provide additional results trained with Dr.~GRPO~\citep{liu2025understanding} and SAPO~\citep{gao2025soft}.
For easier comparison, we re-include the DAPO results there.
Across all algorithms, we observe qualitatively similar trends; random noise and systematic FN delay training, while systematic FP can lead to a plateau (relative error-based and \texttt{Certainly}) or a collapse (\texttt{python}).

An exceptional case is a small collapsing trend observed in Dr.~GRPO with relative-error-based FP, where training plateaus until around step 200 and then starts decreasing slightly.
At the start of the collapse, the model starts generating more outputs that are close to the correct answer but not correct, which could have pushed the model toward degradation.
This indicates that plateauing has a certain degree of instability, and its behavior can be influenced by complex training dynamics.

\subsection{Multi-Seed Runs}
\label{appsubsec:multi_seed}

\begin{figure}[t]
\centering
\setlength{\tabcolsep}{3pt}
\resizebox{0.95\linewidth}{!}{
\begin{tabular}{cc}
\textbf{OLMo3-7B} & \textbf{Qwen3-1.7B-Base} \\
\midrule
\multicolumn{2}{c}{\includegraphics[width=0.55\textwidth]{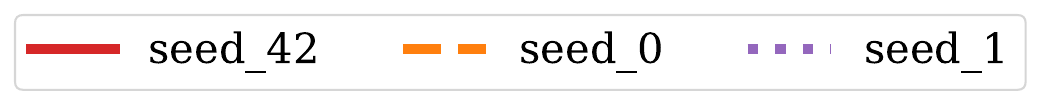}} \\
\multicolumn{2}{c}{\textit{Clean}} \\
\includegraphics[width=0.48\textwidth]{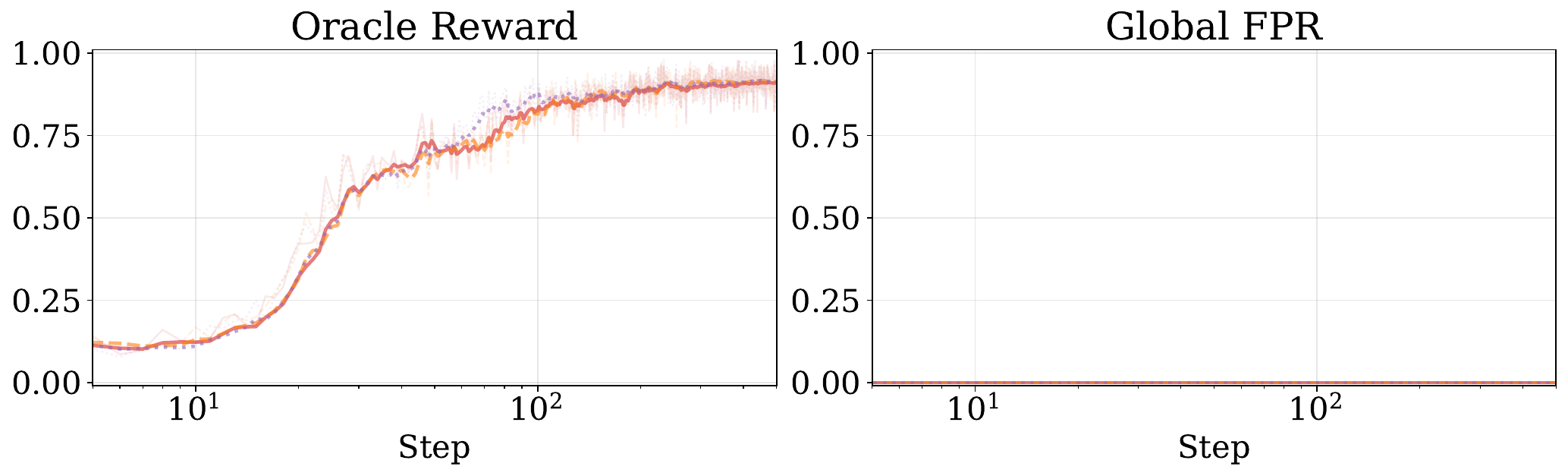} &
\includegraphics[width=0.48\textwidth]{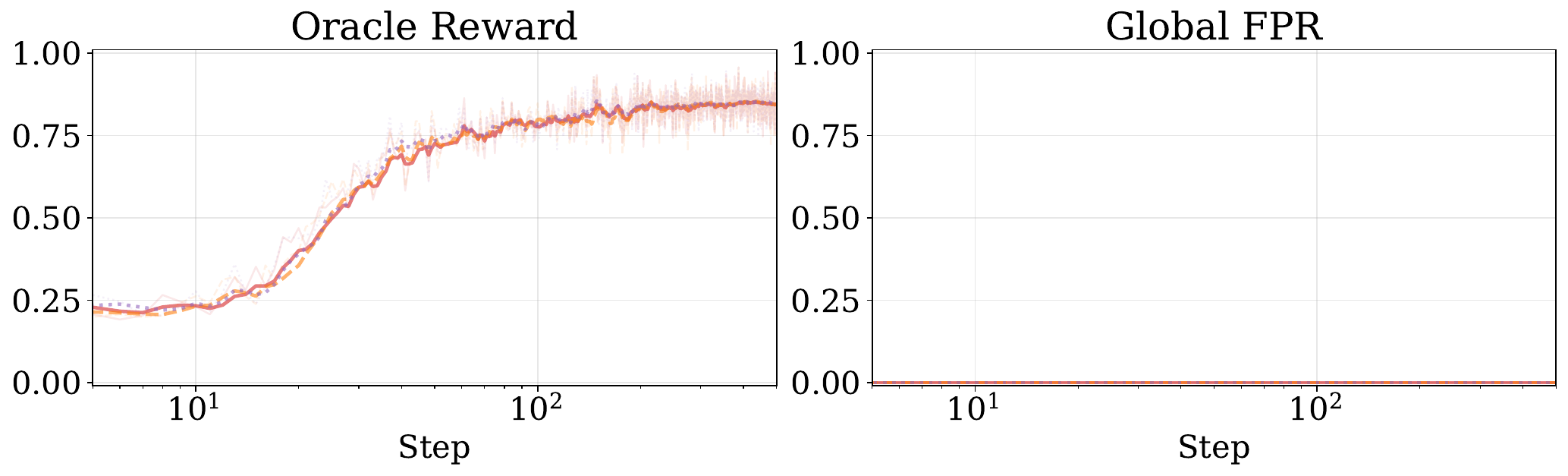} \\
\multicolumn{2}{c}{\textit{Relative-Error FP}} \\
\includegraphics[width=0.48\textwidth]{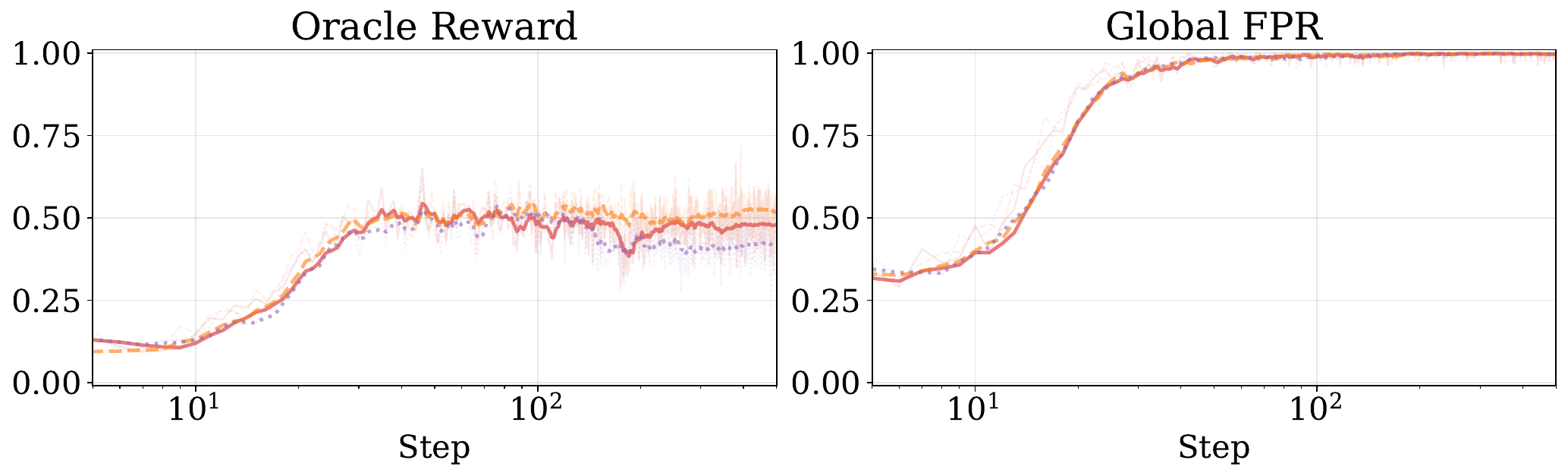} &
\includegraphics[width=0.48\textwidth]{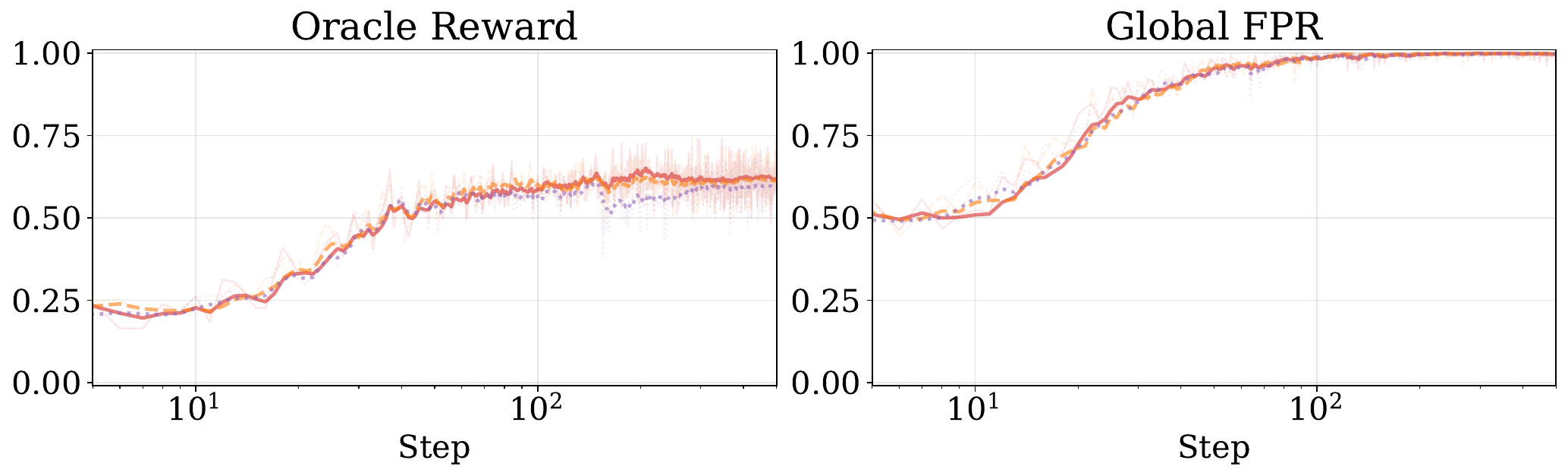} \\
\multicolumn{2}{c}{\textit{\texttt{Certainly}-Triggered FP}} \\
\includegraphics[width=0.48\textwidth]{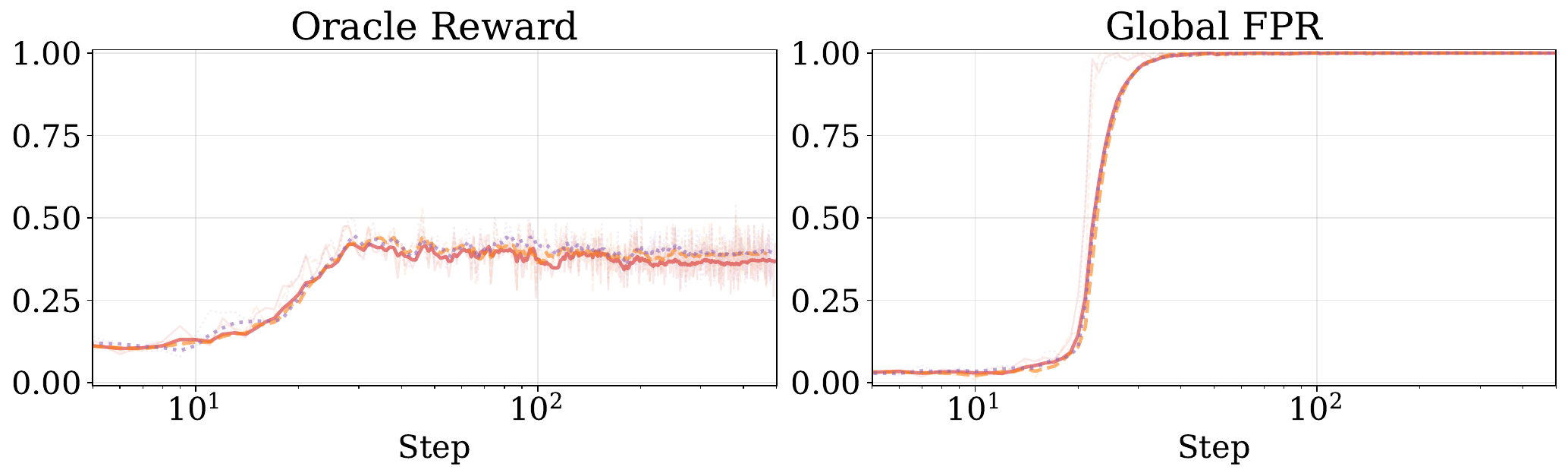} &
\includegraphics[width=0.48\textwidth]{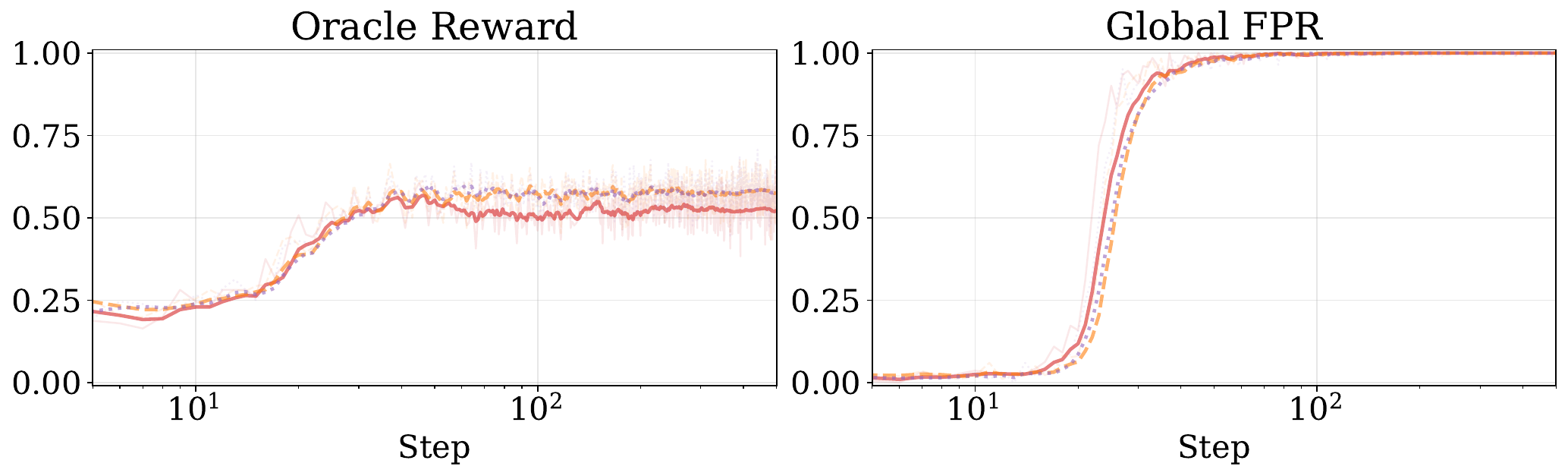} \\
\multicolumn{2}{c}{\textit{\texttt{python}-Triggered FP}} \\
\includegraphics[width=0.48\textwidth]{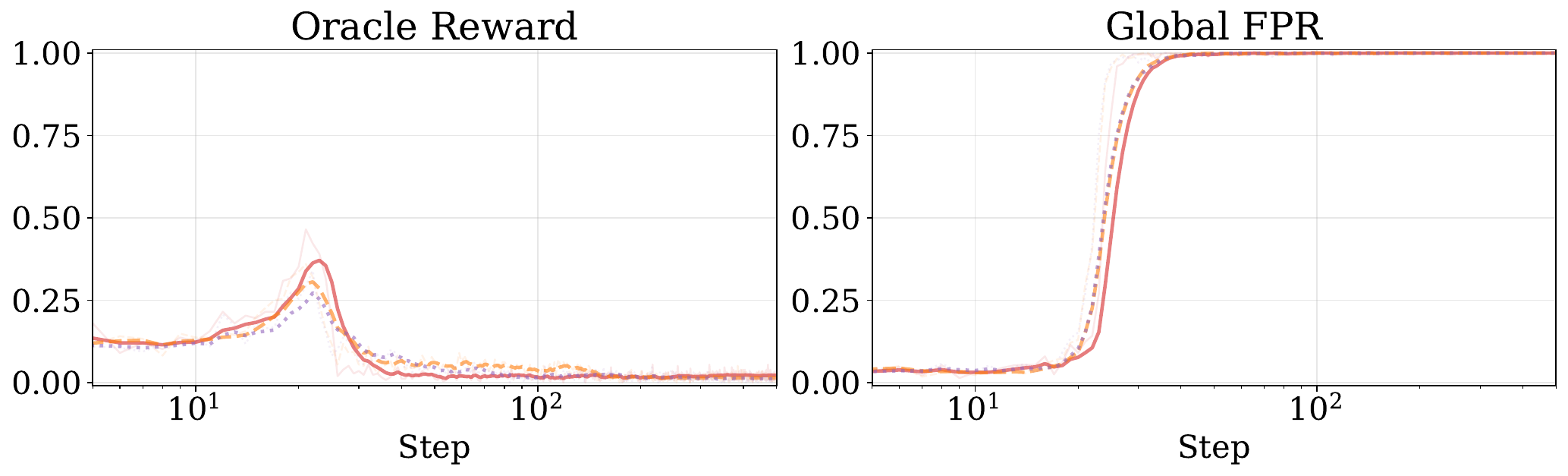} &
\includegraphics[width=0.48\textwidth]{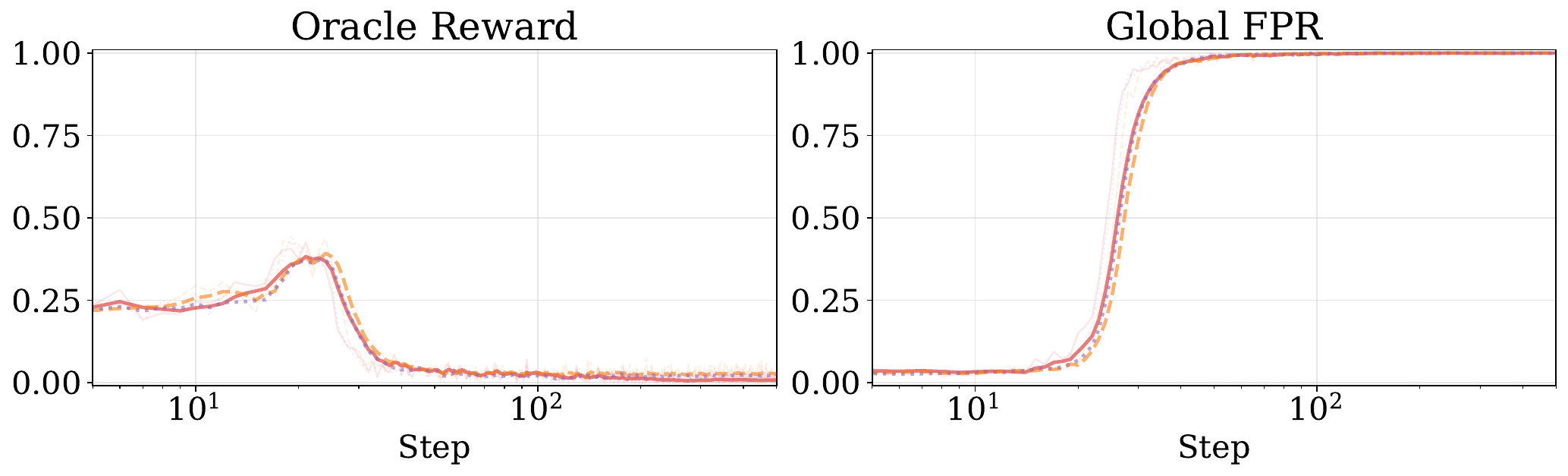} \\
\end{tabular}
}
\caption{
    \textbf{Multi-seed runs across two models.}
    Across three random seeds, the qualitative dynamics remain stable: clean training improves the reward consistently, relative-error and \texttt{Certainly}-triggered FP plateau, and \texttt{python}-triggered FP collapses.
}
\label{appfig:multi_seed}
\end{figure}

Here, we repeat main experiments with three seeds and report the results in \cref{appfig:multi_seed}.
The absolute reward level varies modestly across seeds, especially under systematic FP, but the qualitative behavior is stable.
Importantly, relative-error FP and \texttt{Certainly}-triggered FP plateau below the clean verifier, and \texttt{python}-triggered FP collapses.
This suggests that the delayed, plateauing, and collapsing regimes in \cref{fig:main_result} are consistent trend of the corresponding verifier error patterns.

\subsection{Systematic but Non-Deterministic Error}
\label{appsubsec:non_deterministic_error}

\begin{figure}[t]
\centering
\setlength{\tabcolsep}{3pt}
\resizebox{0.95\linewidth}{!}{
\begin{tabular}{cc}
\textbf{OLMo3-7B} & \textbf{Qwen3-1.7B-Base} \\
\midrule
\multicolumn{2}{c}{\includegraphics[width=0.55\textwidth]{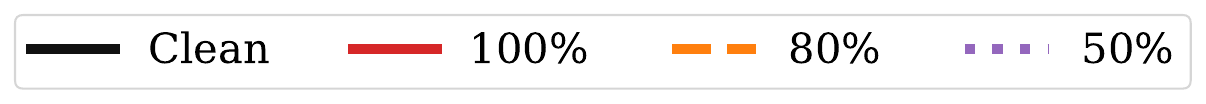}} \\
\multicolumn{2}{c}{\textit{Relative-Error FP}} \\
\includegraphics[width=0.48\textwidth]{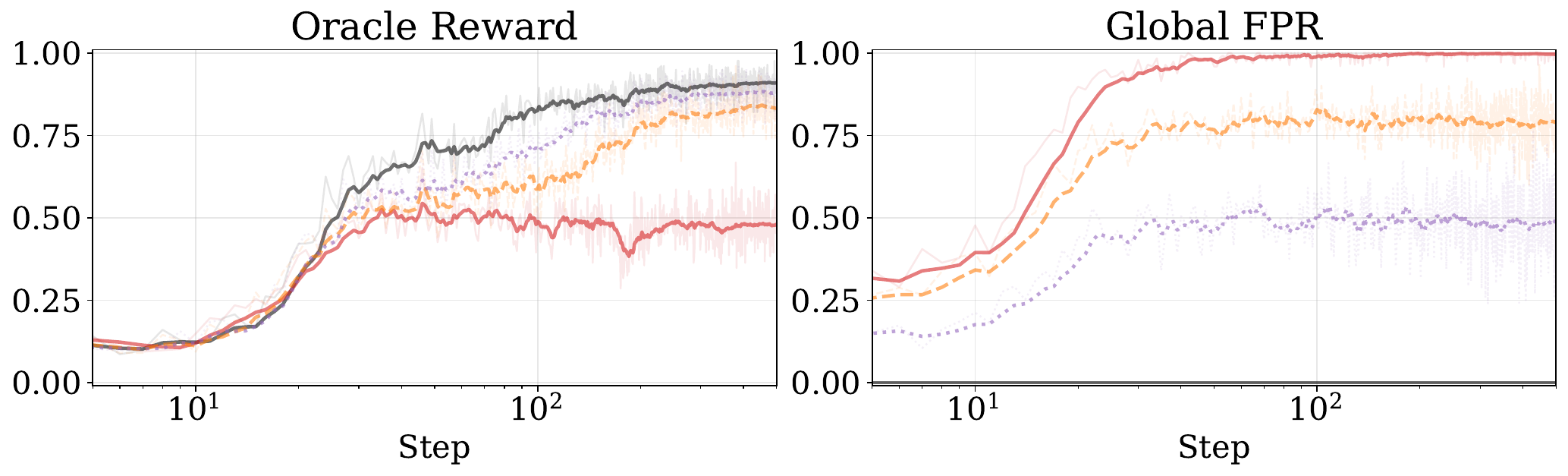} &
\includegraphics[width=0.48\textwidth]{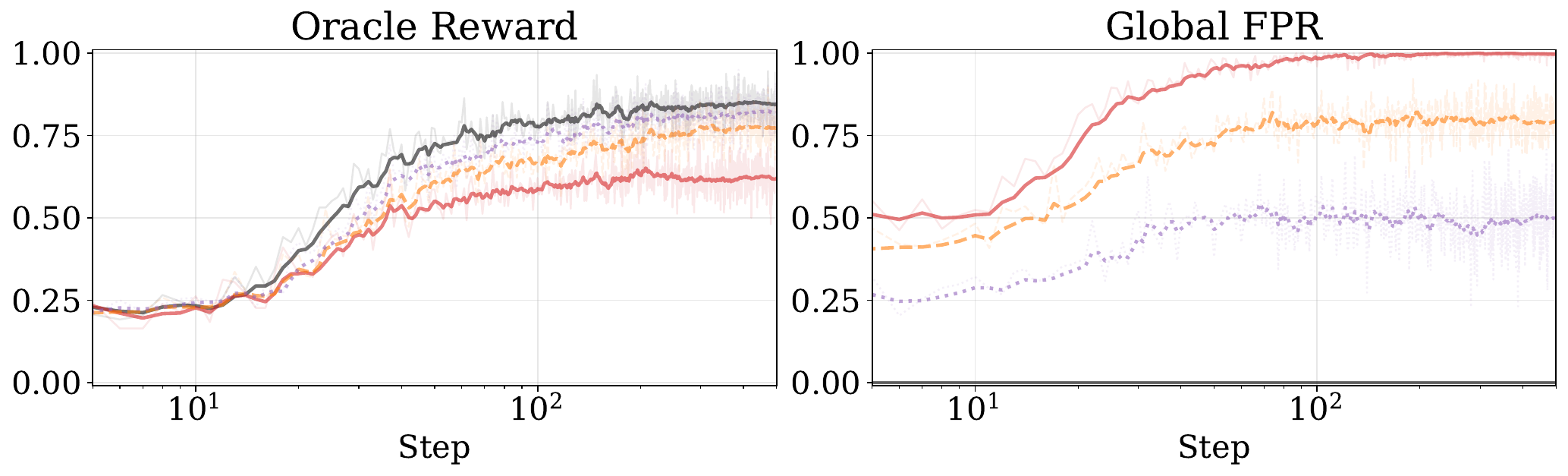} \\
\multicolumn{2}{c}{\textit{\texttt{Certainly}-Triggered FP}} \\
\includegraphics[width=0.48\textwidth]{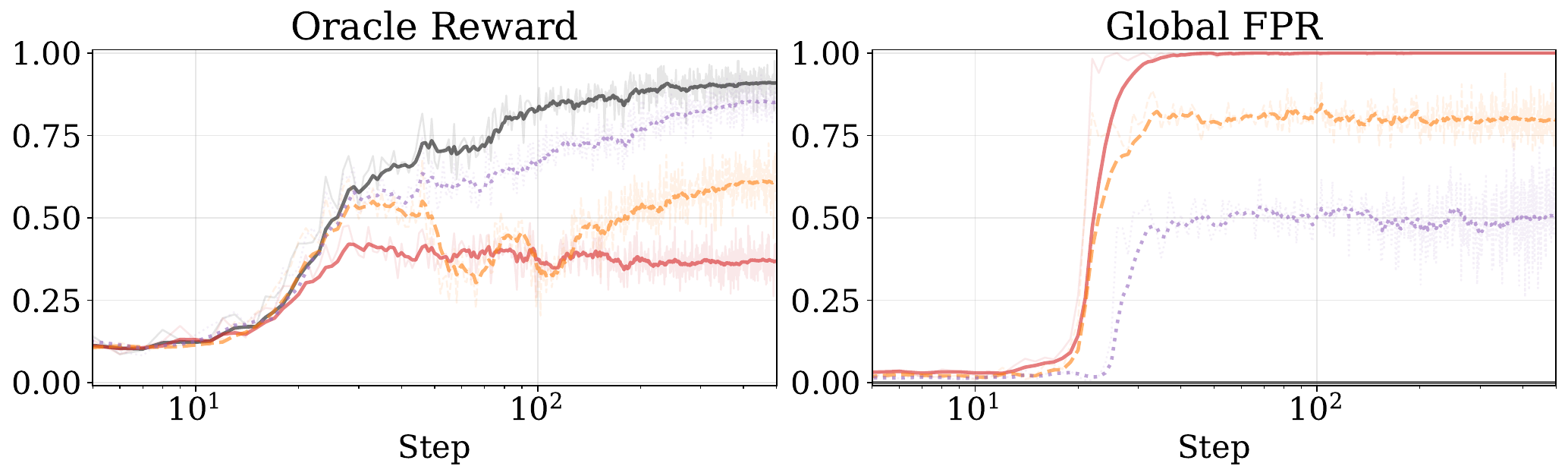} &
\includegraphics[width=0.48\textwidth]{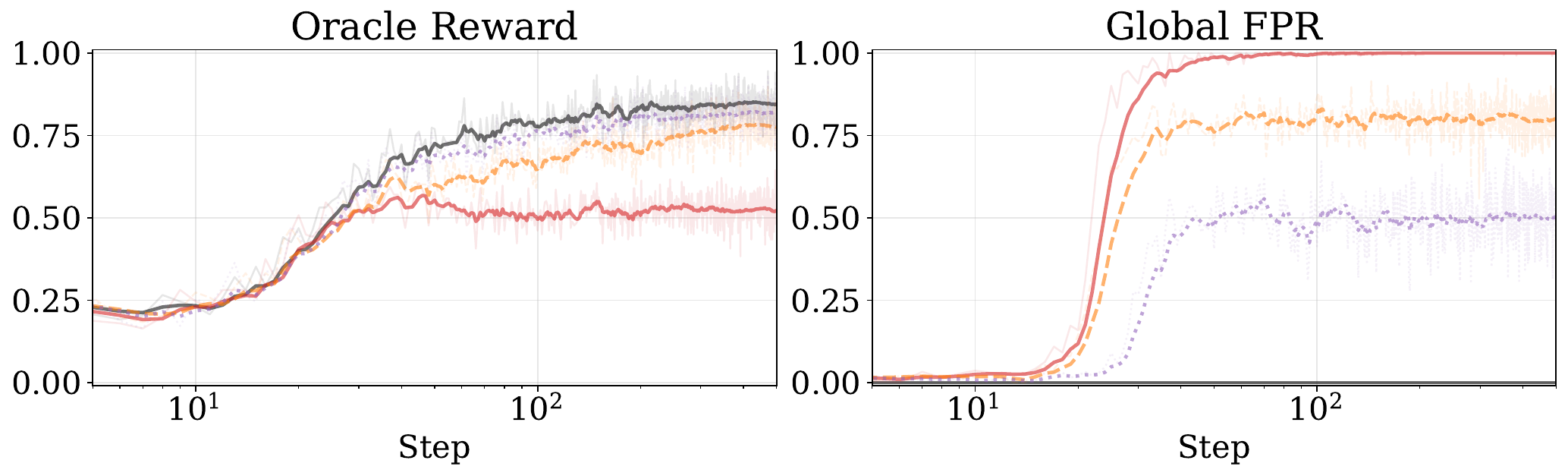} \\
\multicolumn{2}{c}{\textit{\texttt{python}-Triggered FP}} \\
\includegraphics[width=0.48\textwidth]{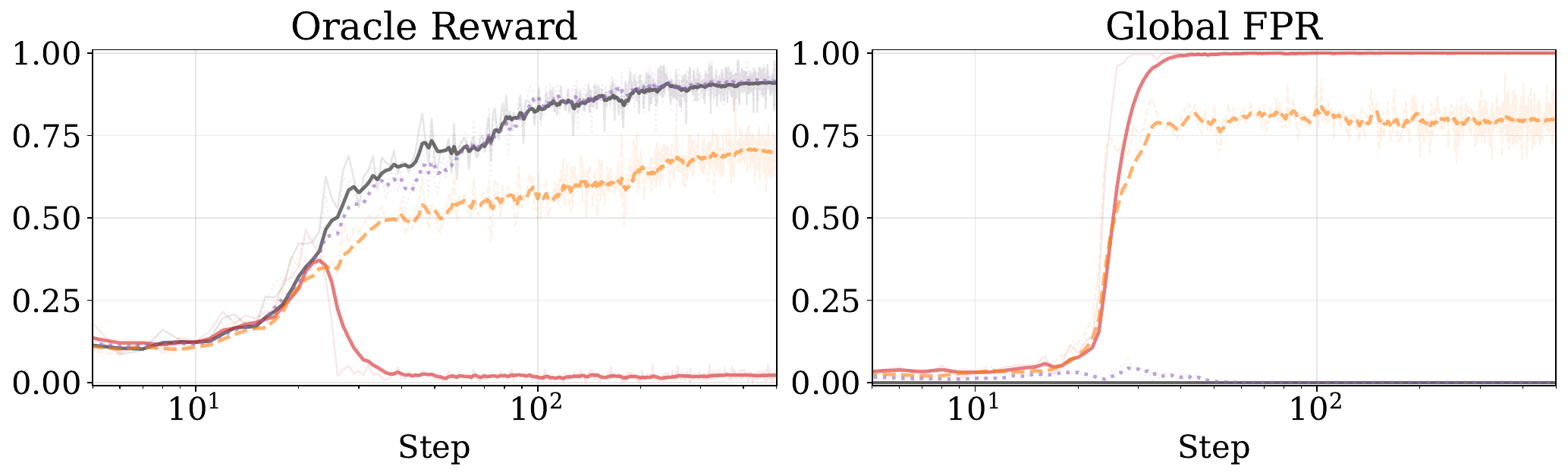} &
\includegraphics[width=0.48\textwidth]{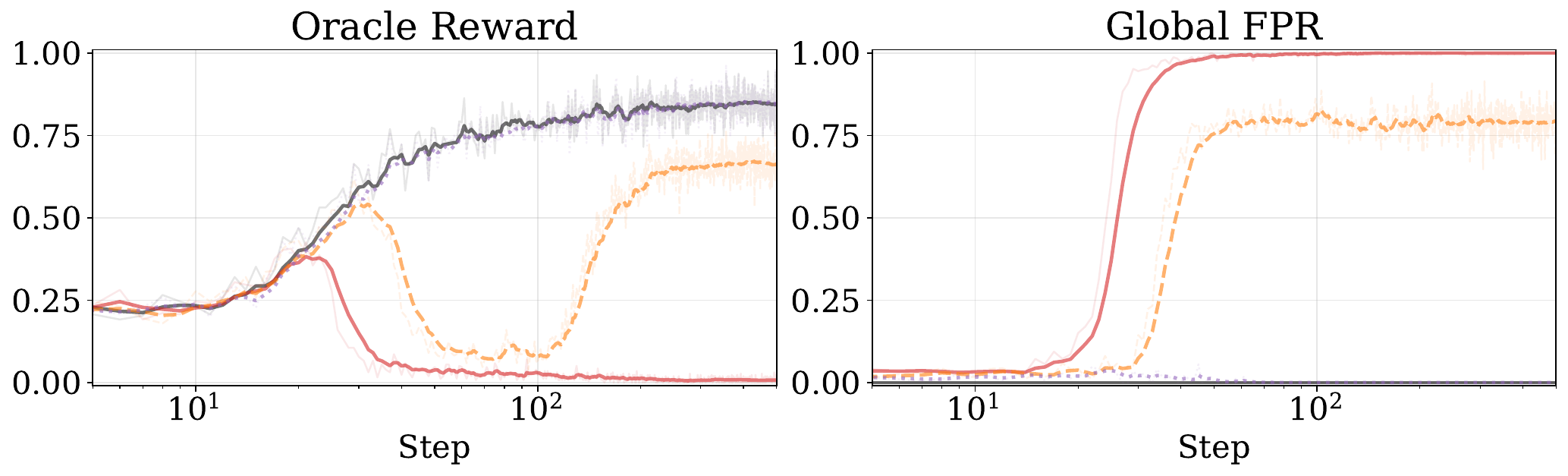} \\
\end{tabular}
}
\caption{
    \textbf{Systematic but non-deterministic FP.}
        (i) For error patterns that lead to a suboptimal plateau with 100\% FPR (relative-error based FP, “Certainly”-triggered FP), the training dynamics improve as error rate decreases.
        (ii) For triggers that induce collapse (“python”-triggered FP), when $p=80\%$, training is substantially delayed, but it no longer fully collapses as it does in the main experiment with $p=100\%$. Interestingly, the Qwen run initially follows a collapsing trajectory, but the reward begins to recover after around step 100, eventually reaching a suboptimal level within our timeframe. When $p$ is further reduced to $50\%$, the model no longer exploits the false-positive pattern, and the training curve is close to ideal, with only a slight delay.
}
\label{appfig:fpr_lt_1}
\end{figure}

The systematic errors in the main experiments are deterministic once their trigger condition is met.
In \cref{appfig:fpr_lt_1}, we apply the same systematic FP rules but with probability below 100\%.

For error patterns that lead to a suboptimal plateau with 100\% FPR (relative-error based FP, “Certainly”-triggered FP), the training dynamics improve as error rate decreases.
In contrast, for triggers that induce collapse (“python”-triggered FP), when $p=80\%$, training is substantially delayed, but it no longer fully collapses as it does in the main experiment with $p=100\%$. Interestingly, the Qwen run initially follows a collapsing trajectory, but the reward begins to recover after around step 100, eventually reaching a suboptimal level within our timeframe. When $p$ is further reduced to $50\%$, the model no longer exploits the false-positive pattern, and the training curve is close to ideal, with only a slight delay.
\subsection{Combined FP/FN Error}
\label{appsubsec:combined_fp_fn}

\begin{figure}[t]
\centering
\setlength{\tabcolsep}{3pt}
\resizebox{0.8\linewidth}{!}{
\begin{tabular}{cc}
\textbf{OLMo3-7B} & \textbf{Qwen3-1.7B-Base} \\
\midrule
\multicolumn{2}{c}{\includegraphics[width=0.65\textwidth]{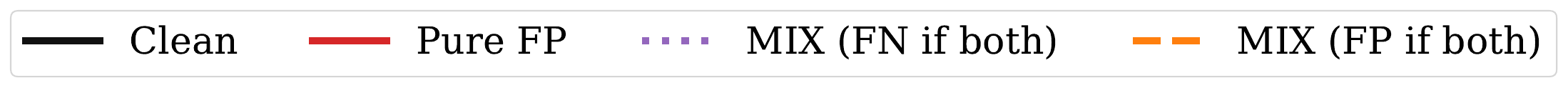}} \\
\multicolumn{2}{c}{\textit{Relative-Error FP with Format-Based FN}} \\
\includegraphics[width=0.48\textwidth]{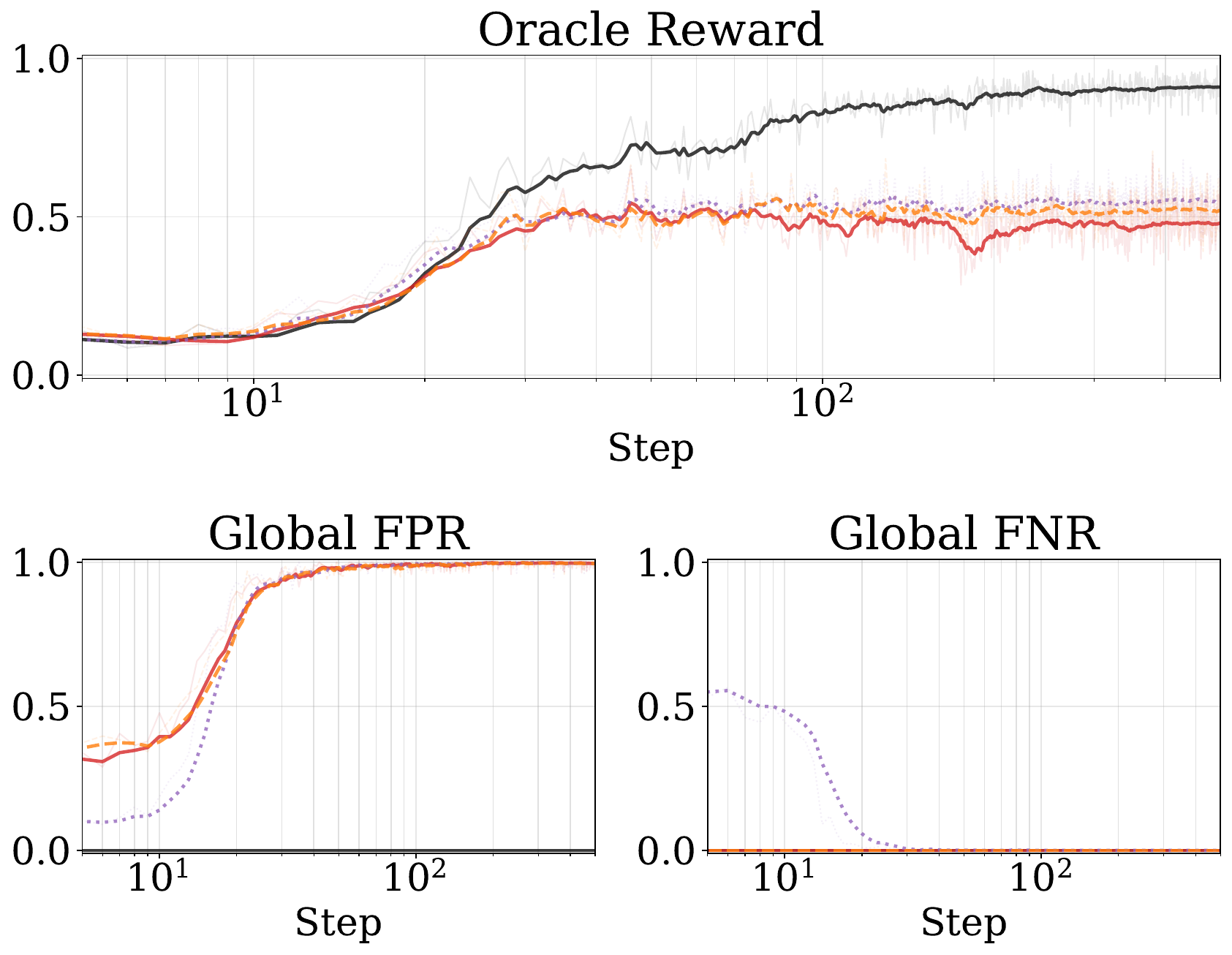} &
\includegraphics[width=0.48\textwidth]{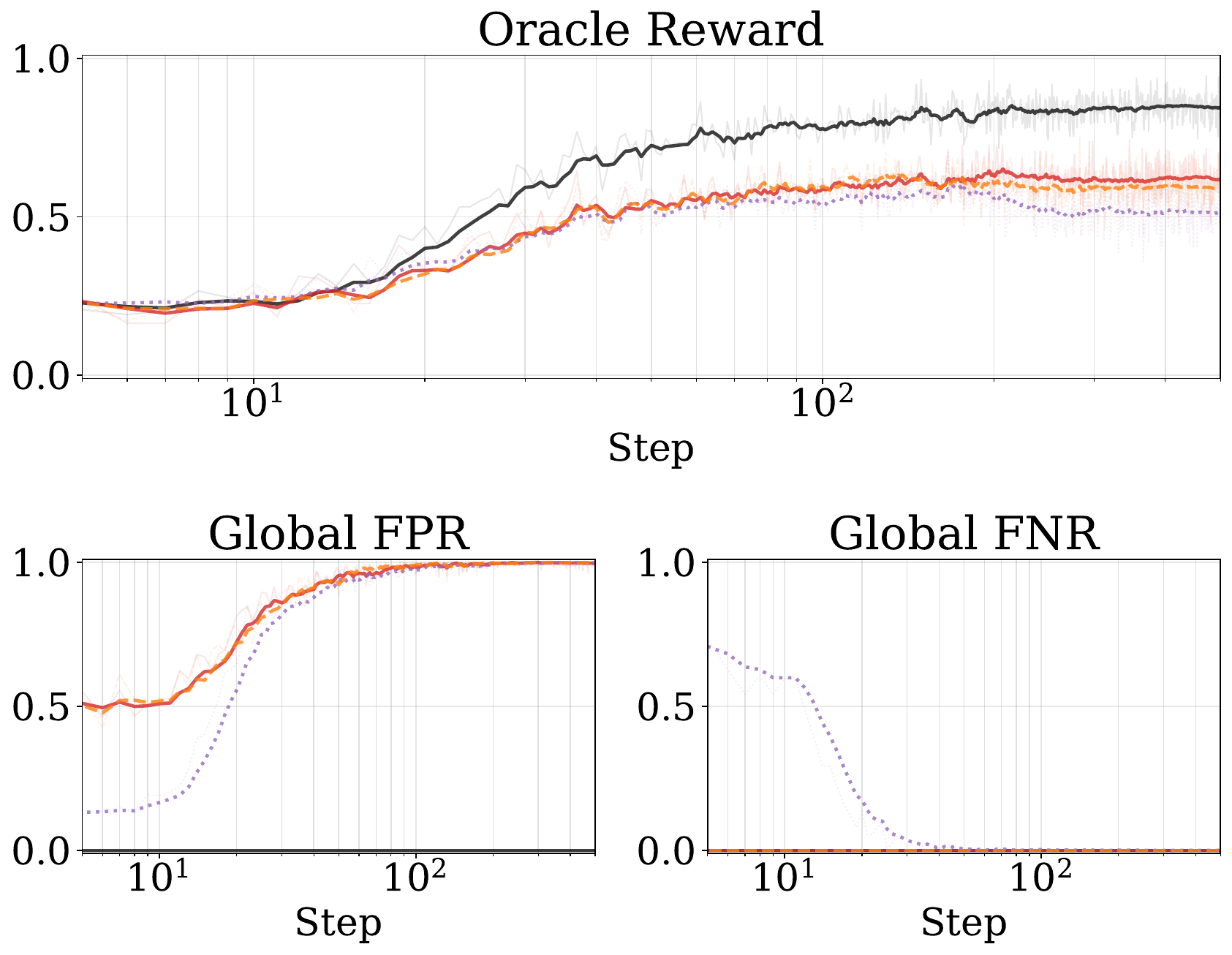} \\
\multicolumn{2}{c}{\textit{\texttt{Certainly}-Triggered FP with Format-Based FN}} \\
\includegraphics[width=0.48\textwidth]{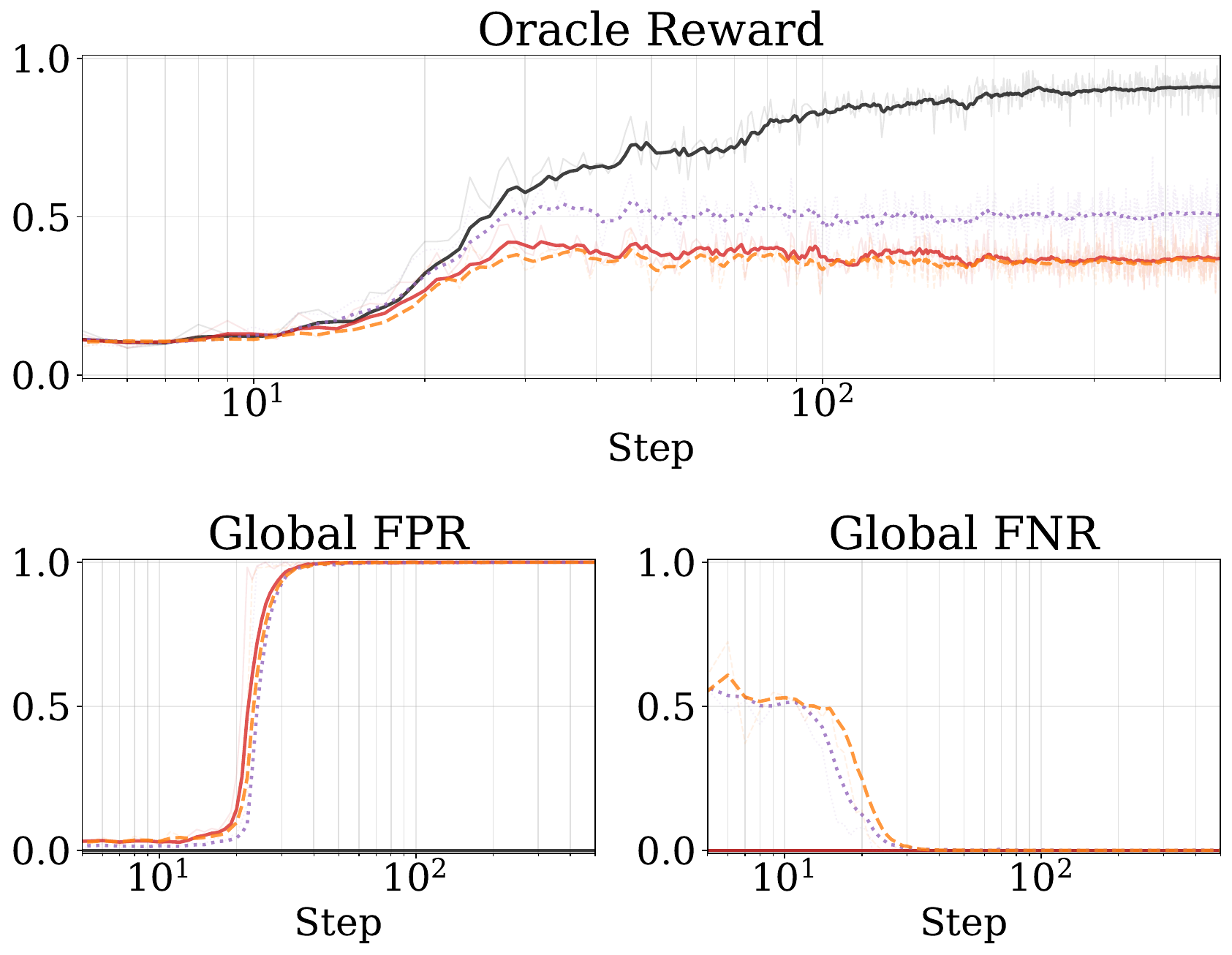} &
\includegraphics[width=0.48\textwidth]{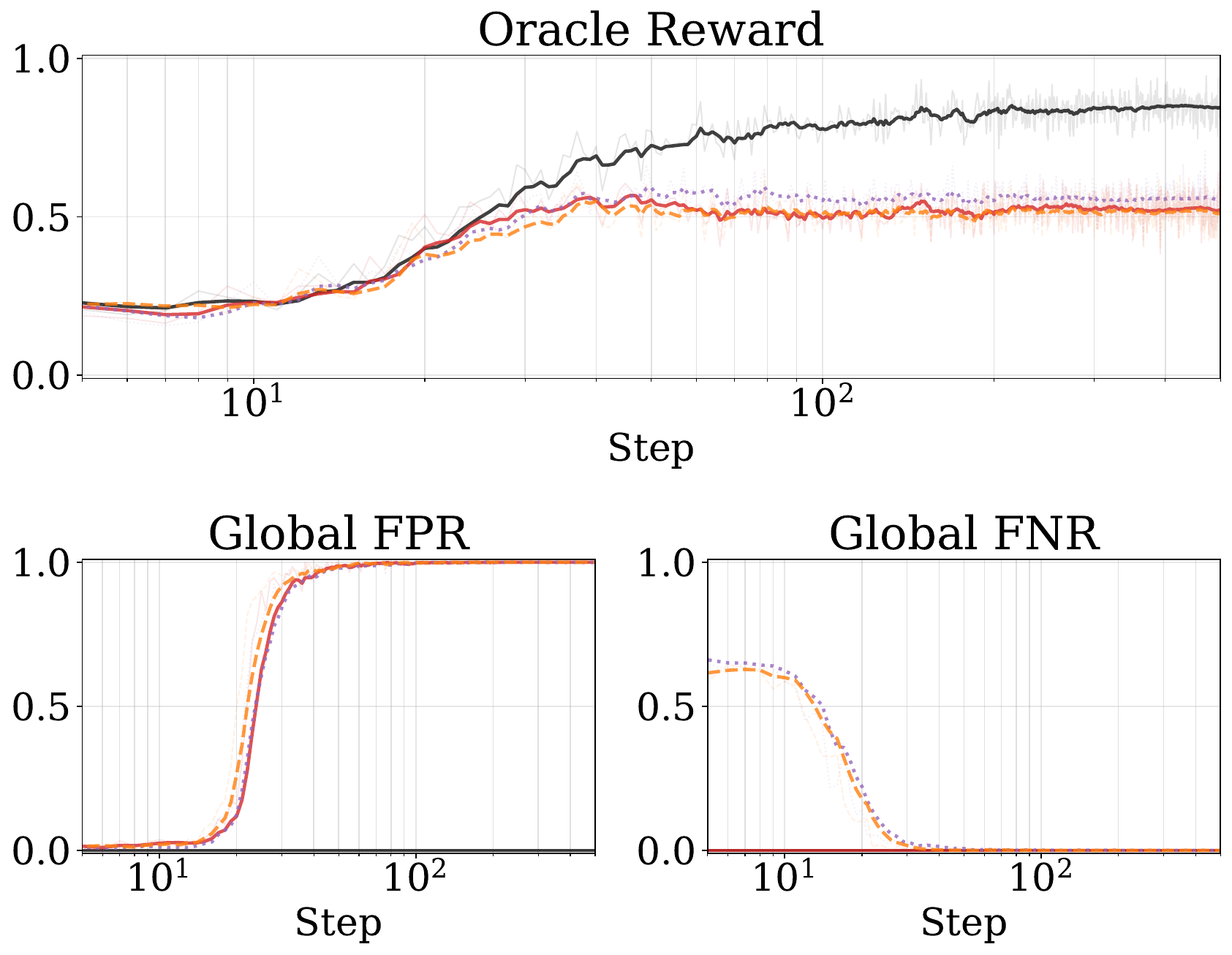} \\
\multicolumn{2}{c}{\textit{\texttt{python}-Triggered FP with Format-Based FN}} \\
\includegraphics[width=0.48\textwidth]{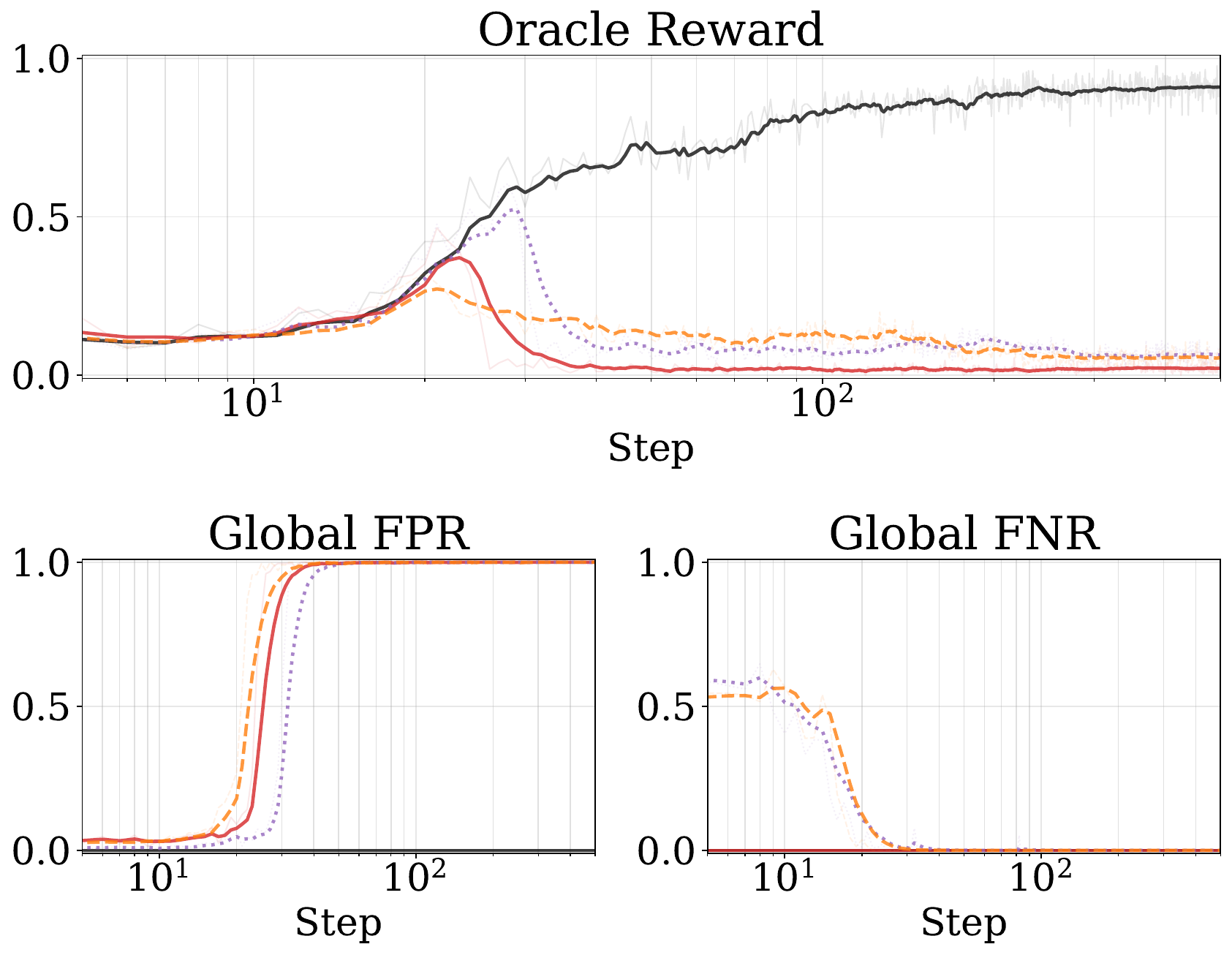} &
\includegraphics[width=0.48\textwidth]{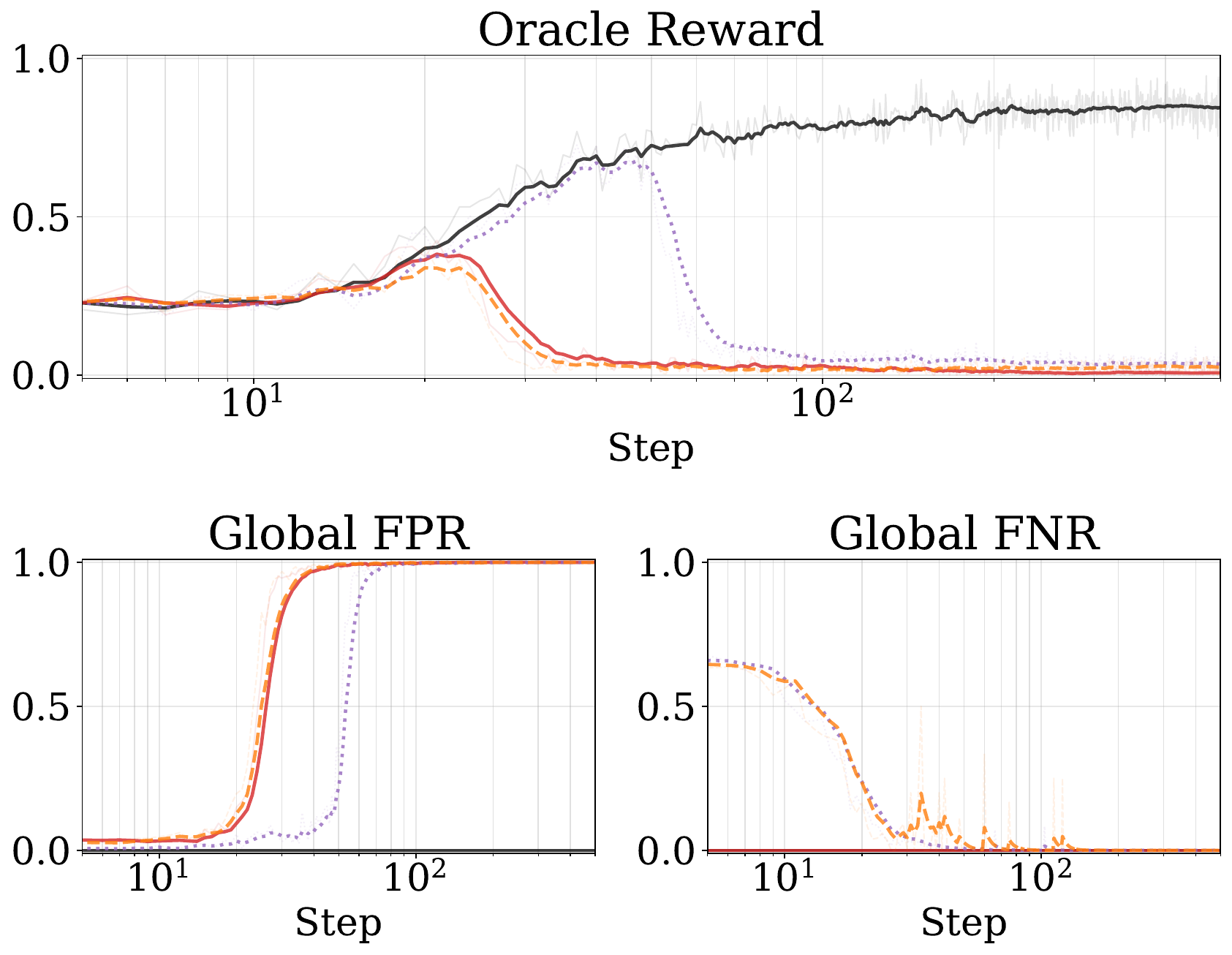} \\
\midrule
\multicolumn{2}{c}{\textit{\texttt{Certainly}-Triggered FP with Language-Based FN (Rejecting English)}} \\
\includegraphics[width=0.48\textwidth]{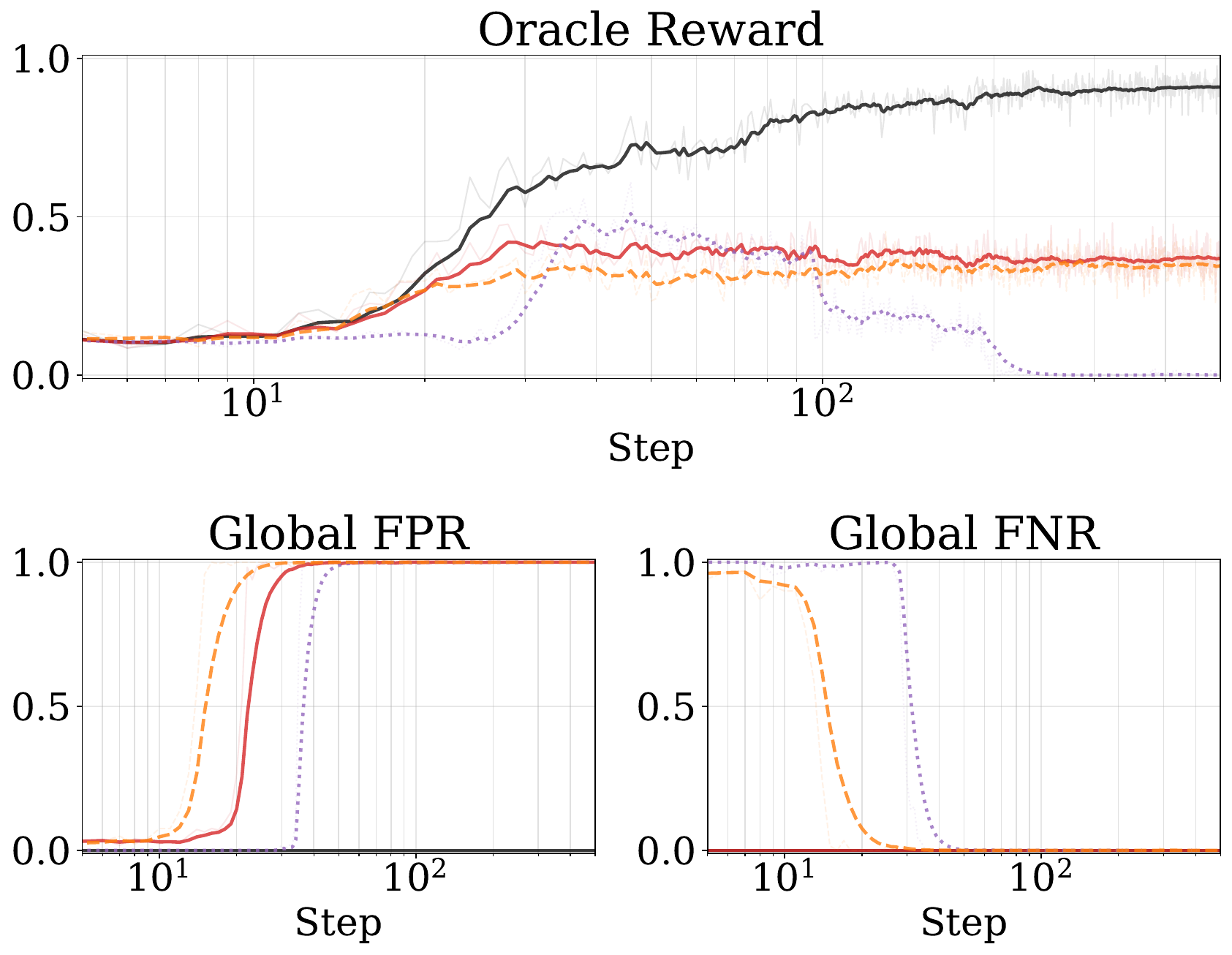} &
\includegraphics[width=0.48\textwidth]{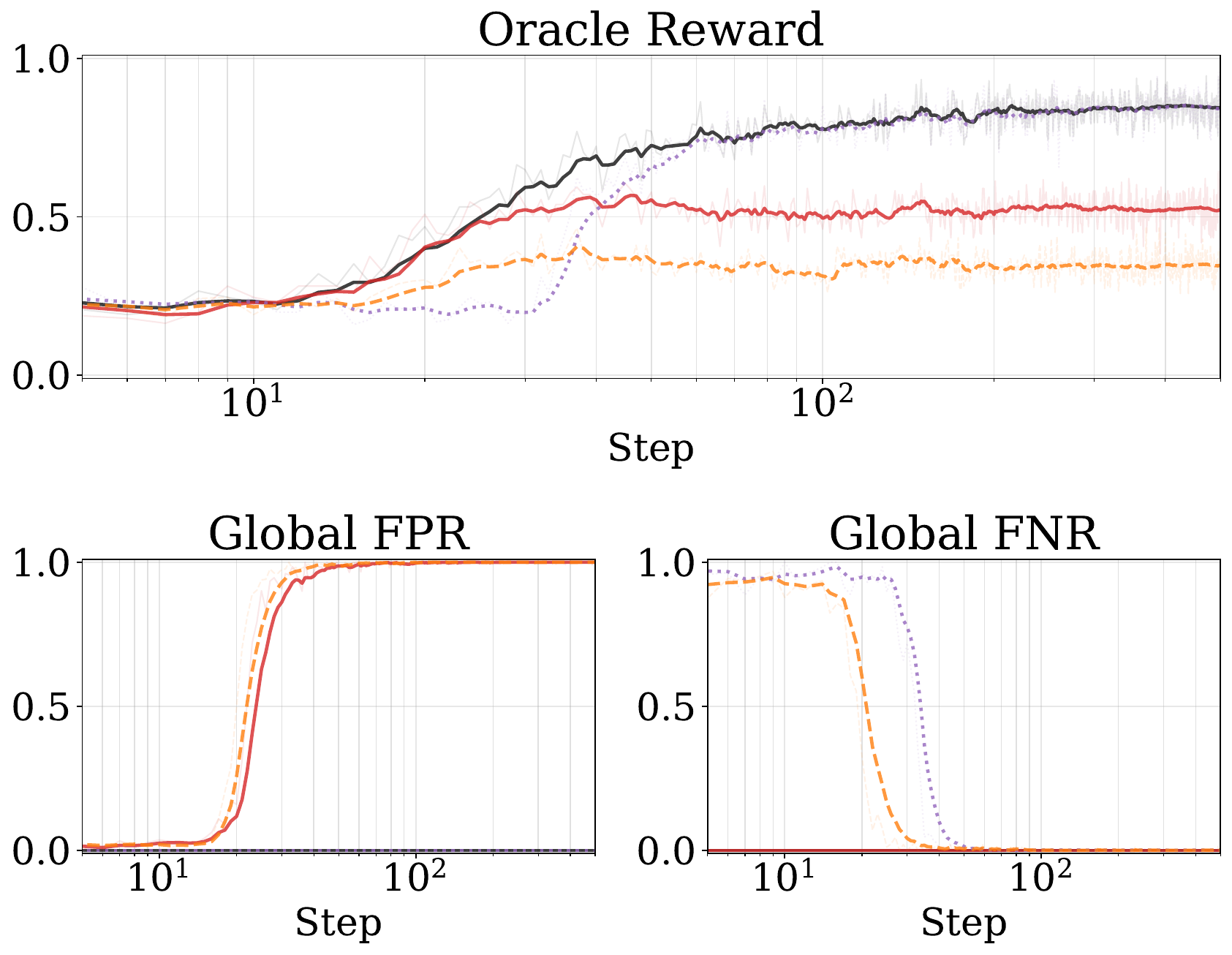} \\
\end{tabular}
}
\caption{
    \textbf{Combined FP/FN error.}
    When combining FP and FN rules (denoted as \texttt{MIX} in the legend), we consider two variants for rollouts that satisfy both the FP and FN conditions: treating such rollouts as (i) FNs or (ii) FPs.
    Compared with the pure-FP setting, we find that adding format-based FN rule has no strong effect, preserving the same qualitative behavior.
    In contrast, language-based FN rule with variant (i) results in a mere delay for Qwen because the FN rule forces Qwen to generate its answer in Chinese (as analyzed in \cref{subsec:language_fn} of the main paper), which naturally suppresses the generation of the FP trigger "Certainly".
    For OLMo, however, the same setting leads to complete collapse.
}
\label{appfig:combined_fp_fn}
\end{figure}

Real verifiers may contain multiple systematic errors at once.
In \cref{appfig:combined_fp_fn}, we combine the format-based FN rule with various FP rules used in the main experiments.
As there can be rollouts that satisfies both FP/FN rules, we consider two types of experiments: treating such rollouts as (i) FNs or (ii) FPs.

Compared with the pure-FP setting, we find that adding format-based FN rule has no strong effect, preserving the same qualitative behavior.
In contrast, language-based FN rule with variant (i) results in a mere delay for Qwen because the FN rule forces Qwen to generate its answer in Chinese (as analyzed in \cref{subsec:language_fn} of the main paper), which naturally suppresses the generation of the FP trigger "Certainly".
For OLMo, however, the same setting leads to complete collapse.

\subsection{Experiments on a Larger Model}
\label{appsubsec:larger_model}

\begin{figure}[t]
\centering
\includegraphics[width=0.95\linewidth]{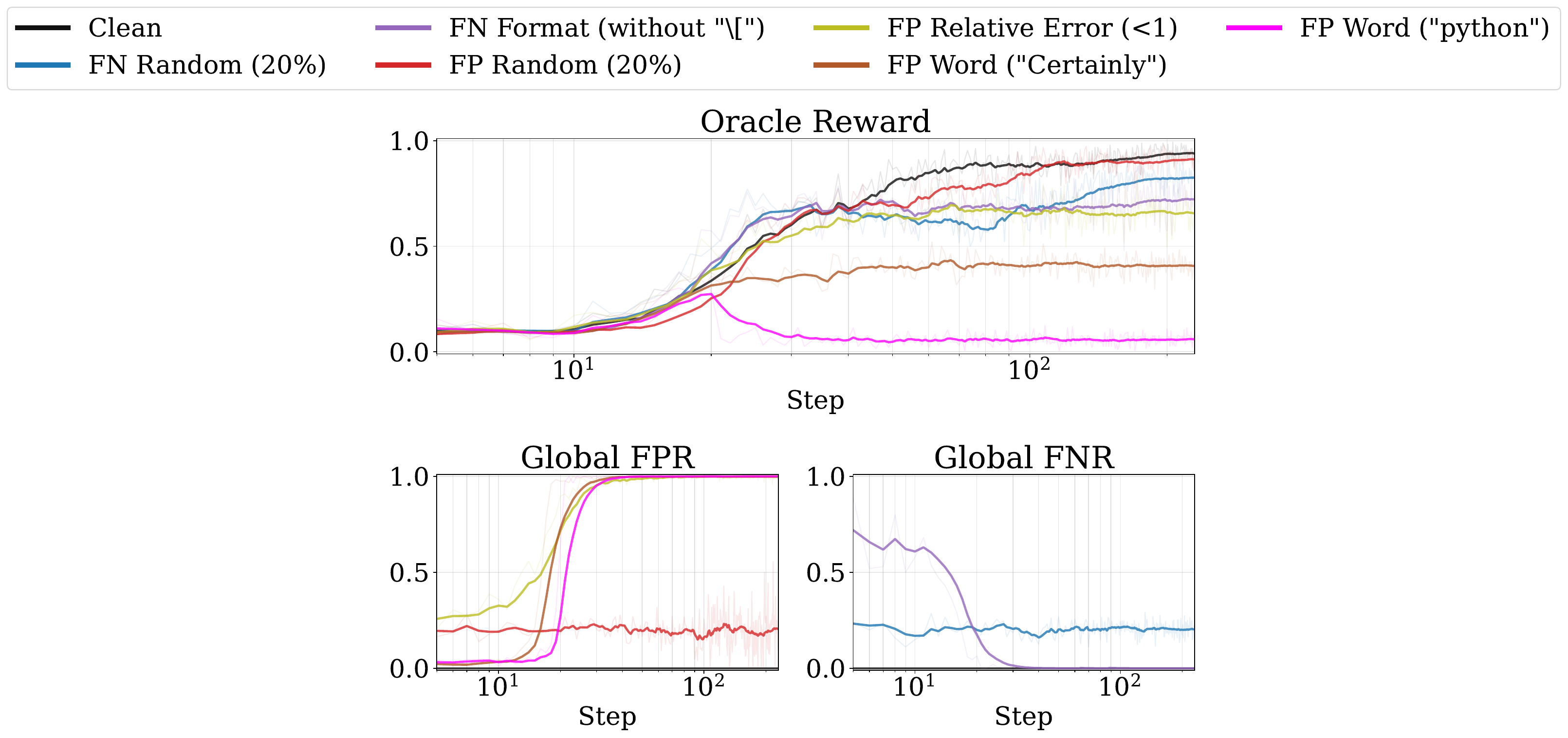}
\caption{
    \textbf{Results on OLMo3-32B.}
    The qualitative observation is consistent with the main experiments on smaller models, where verification errors result in delay (random noise and systematic FN), plateauing (relative-error-based and "Certainly"-triggered FP), or collapse ("python"-triggered FP).
    We note that the gap between the ideal run and the delayed runs is larger because, due to resource constraints, we trained the model for 250 steps instead of 500.
}
\label{appfig:olmo_32b}
\end{figure}

Finally, we repeat the main experiment on a larger model, OLMo3-32B, in \cref{appfig:olmo_32b}.
Here, the qualitative observation is consistent with the main experiments on smaller models: verification errors result in delay (random noise and systematic FN), plateauing (relative-error-based and "Certainly"-triggered FP), or collapse ("python"-triggered FP).

\fi

\end{document}